\documentclass[conference,10pt]{IEEEtran}

\usepackage{graphicx}
\usepackage{subfigure}
\usepackage{amsmath}
\usepackage{amssymb}
\usepackage{booktabs}
\usepackage{multirow}
\usepackage[square,numbers,sort&compress]{natbib}
\usepackage{setspace}
\usepackage{anyfontsize}
\ifCLASSINFOpdf
\else
\fi
\begin{document}
%
\title{Enhancing Adversarial Attacks on Single-Layer NVM Crossbar-Based Neural Networks with Power Consumption Information\vspace{-2mm}}



%
\author{\IEEEauthorblockN{Cory Merkel}
\IEEEauthorblockA{Brain Lab\\
Rochester Institute of Technology,
Rochester, NY 14623\\ Email: cemeec@rit.edu}
\vspace{-8mm}}


\vspace{-5mm}

\maketitle
\normalsize
\begin{abstract}
Adversarial attacks on state-of-the-art machine learning models pose a significant threat to the safety and security of mission-critical autonomous systems.  This paper considers the additional vulnerability of machine learning models when attackers can measure the power consumption of their underlying hardware platform.  In particular, we explore the utility of power consumption information for adversarial attacks on non-volatile memory crossbar-based single-layer neural networks.  Our results from experiments with MNIST and CIFAR-10 datasets show that power consumption can reveal important information about the neural network's weight matrix, such as the 1-norm of its columns.  That information can be used to infer the sensitivity of the network's loss with respect to different inputs.  We also find that surrogate-based black box attacks that utilize crossbar power information can lead to improved attack efficiency.
\end{abstract}


%
\IEEEpeerreviewmaketitle

\section{Introduction}

Advancements in machine learning (ML) are opening up new domains for artificial intelligence (AI) at record speed.  In order to keep pace, hardware designers have also been developing new and innovative methods to accelerate and improve the efficiency of ML algorithms \cite{bavikadi2022survey}.  One of the leading ideas in this space is to closely couple computation and memory, leading to ``in-memory" computing, where the memory array itself actively participates in computation \cite{sebastian2020memory}.  In fact, computing with passive arrays of non-volatile memory (NVM) has become one of the most popular motifs in the literature for the design of size, weight, and power-efficient AI accelerators.  In particular, dense arrays of NVM devices are able to efficiently perform matrix-vector multiplication, which is ubiquitous in state-of-the-art ML algorithms like deep neural networks.

As our dependence on AI grows, it is critical to study the unique interplay between ML algorithms and new hardware platforms to determine when their decisions should (and should not) be trusted.  Recently, the robustness, security, and safety of ML algorithms like deep neural networks has been called into question through research in adversarial machine learning (AML).   AML concerns both the offensive and defensive measures associated with malperformance and/or privacy of ML.  There are several types of AML attacks and defenses \cite{joseph2018adversarial,vorobeychik2018adversarial,biggio2018wild} that have been proposed in the literature.  Evasion attacks, which are the most popular, were introduced in the mid-2000's, as small perturbations to the content of emails, causing them to be misclassified by linear spam filters \cite{dalvi2004adversarial,lowd2005adversarial,lowd2005good}.  In 2014, Szegedy et al. \cite{szegedy2013intriguing} showed that these types of attacks can also be performed on deep learning models, showing that imperceptible perturbations in the pixel space of images led to high-confidence misclassifications by convolutional neural networks.  Since then, there has been an explosion in research on AML, especially in the context of deep learning \cite{biggio2018wild}.  However, the implications for AML in the context of emerging NVM-based hardware platforms still needs deeper exploration.  Previous work in this area is relatively sparse.  Roy et al. \cite{roy2021intrinsic} as well as conceptually similar works \cite{cherupally2021leveraging,bhattacharjee2021neat} showed that non-idealities in NVM crossbars help to mitigate adversarial attacks if they can not be accurately modeled.  Moitra et al. \cite{moitra2021detectx} explored the idea of using current signatures in NVM crossbars to detect adversarial attacks.  In \cite{barve2021adversarial}, the authors explore the adversarial robustness of reconfigurable NVM crossbar-based accelerators trained with a range of algorithms and assuming different device characteristics.

In this paper, we explore NVM crossbar-based neural networks' vulnerability to adversarial attacks when the attacker is able to measure their power consumption.  To the best of our knowledge, this is the first work exploring the use of power consumption information for crafting adversarial attacks against NVM crossbar-based ML accelerators.  We believe that the results shown here represent an important contribution to the existing body of work in this area and give motivation for further research on power analysis for adversarial attacks against this class of hardware.  The rest of this paper proceeds as follows:  Section \ref{section:background} provides background on adversarial evasion attacks and NVM crossbars.  Sections \ref{section:nooutput} discusses results for the case where the attacker only has access to power consumption information and cannot observe the neural network outputs.  Section \ref{section:output} presents results for the case where the attacker does have access to the neural network outputs.  Finally, Section \ref{section:conclusion} concludes this work.

\section{Background}
\label{section:background}

\subsection{Adversarial Evasion Attacks}
\begin{figure}[!t]
\centering
\subfigure[]{
\includegraphics[width=0.25\textwidth]{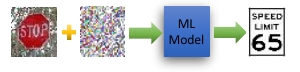}
\label{fig:stop_sign}
}
\hspace{-6mm}
\subfigure[]{
\includegraphics[width=0.23\textwidth]{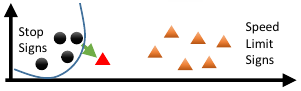}
\label{fig:evasion_attack}
}
\caption{(a) High-level description of an evasion attack showing a stop sign misclassified as a speed limit sign.  (b) Evasion attacks work by moving datapoints in the input space across the model's decision boundary.}
\label{fig:attacks}
\vspace{-6mm}
\end{figure}

The idea behind an adversarial evasion attack is illustrated in Figure \ref{fig:stop_sign}.  Here, the input to a machine learning model, an image in this case, is perturbed by a small amount (usually imperceptible to humans), causing a misclassification by the model.  An explanation for this behavior is shown in Figure \ref{fig:evasion_attack}.  Evasion attacks work by exploiting small-margin decision boundaries (points in the input space where the model's label changes).  Often, when training is complete, the model's decision boundaries will lie close enough to the training and test data that small perturbations can cause misclassification.

The goal of an evasion attack can be expressed as an optimization problem, where, for some model $\Pi$, a correctly-classified input $\mathbf{u}$ is perturbed by $\mathbf{r}^{*}$ to maximize a loss function $\mathcal{L}$ and cause $\Pi$'s classification of $\mathbf{u}^{\prime}=\mathbf{u}+\mathbf{r}^{*}$ to be different from $\mathbf{u}$'s ground truth label:

\begin{equation}
\begin{aligned}
\mathbf{r}^{*}= \underset{\mathbf{r}\in\mathcal{R},}{\arg\max}&\quad \mathcal{L}(\mathbf{u}+\mathbf{r},l)\\
\textrm{s.t.}&\quad l^{\prime}\ne l
\label{eqn:evasion}
\end{aligned}
\end{equation}

\noindent where $l$ is $\mathbf{u}$'s ground truth label, $l^{\prime}$ is the model's label for $\mathbf{u}^{\prime}$, and $\mathcal{R}$ is the set of set of allowed perturbations.  $\mathbf{u}^{\prime}$ is called an adversarial example.  Often, the allowed set of perturbations takes the form of an $\ell^{p}$-norm constraint: $\mathcal{R}=\{\mathbf{r}\in\mathbb{R}^{N}:\:||\mathbf{r}||_{p}\le D\}$ where $N$ is the dimension of the input space. The $\ell^{p}$-norm of $\mathbf{r}$ is usually bounded in a way that the difference between  $\mathbf{u}$ and $\mathbf{u}^{\prime}$ is difficult or impossible to perceive by a human.  Attacks are usually performed using $\ell^{p}$-norms with $p=2$ or $p=\infty$.  However, $p=0$ and $p=1$ are also common.  $\mathcal{R}$ may also be formed using multiple $\ell^{p}$-norms, box constraints (e.g. bounding all inputs between a minimum and maximum value), or by choosing $\mathbf{r}$ as some type of transformation that imposes a dependence between the elements of $\mathbf{r}$ (e.g. affine transformations such as rotation, scaling, etc.).  A number of evasion attacks have been proposed based on (\ref{eqn:evasion}), differing primarily in the way they define $\mathcal{R}$, how much information they assume is known about $\Pi$ (white box vs. black box attacks), the way they approach the optimization procedure, and whether they are targeted (e.g. classifying a stop sign image as a speed limit sign) or untargeted (e.g. classifying a stop sign image as anything other than a stop sign).  One of the most popular classes of evasion attacks are those based a single step of moving an input in the direction of increasing loss.  These include the fast gradient sign method (FGSM) and fast gradient value (FGV) attacks \cite{goodfellow2014explaining}.  For an FGSM attack:
\begin{equation}
\mathbf{r}^{*}\equiv\epsilon\mathrm{sgn}\left(\nabla_{\mathbf{u}}\mathcal{L}\right)
\end{equation}
\noindent where $\epsilon$ is a scalar constant called the attack strength.
\vspace{-1mm}

\subsection{NVM Crossbar Overview}

\begin{figure}[!t]
    \centering
    \includegraphics[width=0.6\columnwidth]{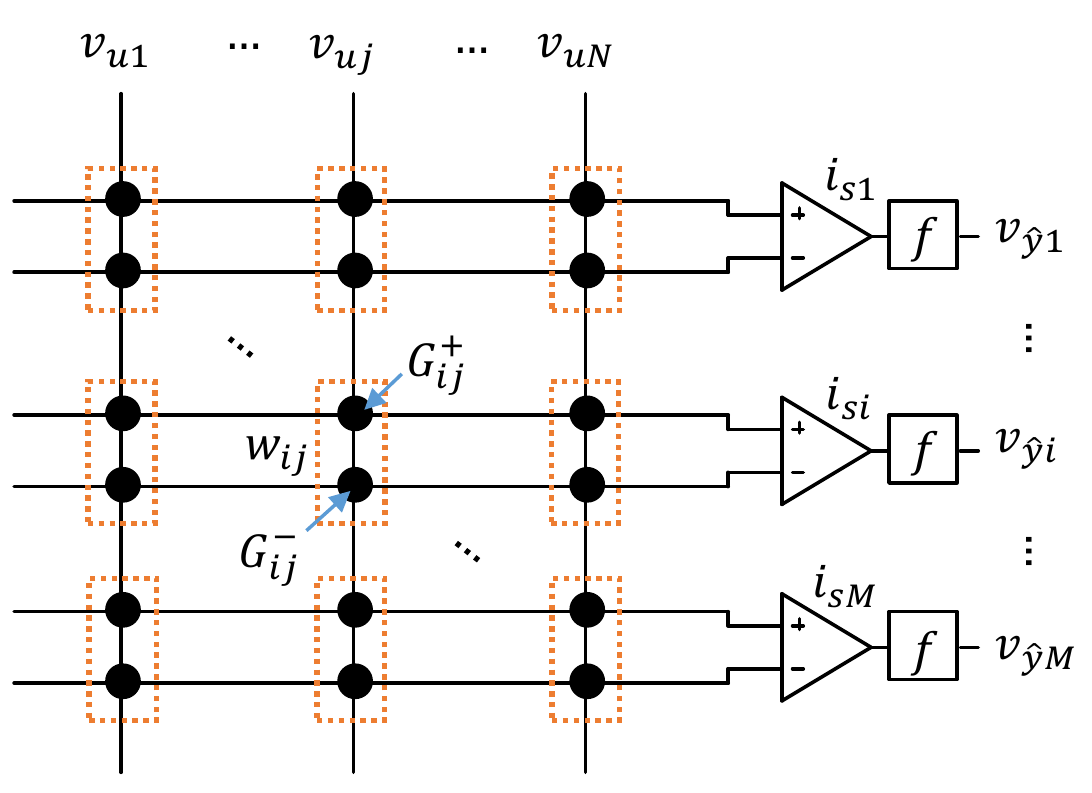}
    \caption{NVM crossbar for implementing a neural network layer.}
    \label{fig:xbar}
    \vspace{-4mm}
\end{figure}

A typical implementation of a neural network layer using an NVM crossbar is shown in Figure \ref{fig:xbar} \cite{zhang2018neuromorphic}.  The layer's weight matrix $\mathbf{W}$ is represented by the conductances of several NVM devices.  There are many choices for the NVM technology that is used, such as resistive random access memory (ReRAM), phase change memory (PCRAM), ferroelectric devices, etc. \cite{upadhyay2019emerging}.  NVM crossbars use Ohm's Law and Kirchoff's Current Law to perform matrix-vector multiplication, which is ubiquitous in neural networks.  The ideal behavior (assuming ohmic NVM devices, no current sneak paths, etc.) of the crossbar is 
\begin{equation}
\mathbf{v}_{\hat{\mathbf{y}}}=f\left(\mathbf{i}_{\mathbf{s}}\right)=f\left(\mathbf{G}\mathbf{v}_{\mathbf{u}}\right)
\label{eqn:xbar}
\end{equation}
where $\mathbf{v}_{\mathbf{u}}$  is the voltage vector of crossbar inputs, $\mathbf{i}_{\mathbf{s}}$ is the current vector of crossbar outputs, and $\mathbf{v}_{\hat{\mathbf{y}}}$ is a voltage vector representing the final layer output after the application of an activation function $f$.  Each element of the conductance matrix $\mathbf{G}$ is represented by two conductances, $G_{ij}\equiv G_{ij}^{+} - G_{ij}^{-}$.  This allows the weights associated with the conductances to either be positive or negative.  The subtraction is represented generically here with a (current-mode) unity gain amplifier, but equivalent behavior could be achieved using other implementations.  By normalizing all of the voltages, currents, and conductances, (\ref{eqn:xbar}) can be rewritten as
\begin{equation}
\hat{\mathbf{y}}=f(\mathbf{s})=f(\mathbf{W}\mathbf{u}).
\end{equation}

Now, the total steady state current flowing through the crossbar can be written as
\begin{equation}
i_{total} = \sum\limits_{j=1}^{N}v_{uj}\sum\limits_{i=1}^{M}G_{ij}^{+}+G_{ij}^{-}= \sum\limits_{j=1}^{N}v_{uj}^{l-1}G_{j},
\end{equation}
where we define $G_{j}$ as the sum of conductances connected to the $j^{\mathrm{th}}$ input.  The unknown values of $G_{j}$ can be determined through several observations of $i_{total}$ for different input voltages.  For example, setting $v_{u1}=V_{dd}$ and grounding all other inputs leads to $G_{1}=i_{total}/V_{dd}$.  The weight value $w_{ij}$ associated with a given pair of neurons $i$ and $j$ is proportional to $G_{ij}^{+}-G_{ij}^{-}$.  This means that there are several ways (potentially infinite) to achieve a given weight with different values of the two conductances.  However, we assume here that for positive weights, $G_{ij}^{-}\approx 0$ and for negative weights, $G_{ij}^{+}\approx 0$.  This is a safe assumption since it will lead to the lowest power consumption for a given weight matrix implementation, which is a primary goal, especially in edge devices.  It also means that each weight will have a one-one mapping to conductance values.  Now, we can write
\begin{equation}
|w_{ij}| \propto G_{ij}^{+}+G_{ij}^{-}
\end{equation}
If $\mathbf{u}$ is known, then by measuring the crossbar current (also referred to in this paper as the power information), we can determine the 1-norm of each column of $\mathbf{W}$.  The utility of this information depends on whether we have access to the crossbar outputs and activation function $f$.

\vspace{-1mm}
\section{Case 1:  Attacker Does Not Have Access to Output}
\label{section:nooutput}
\vspace{-2mm}

\begin{figure*}[!t]
    \centering
    \hspace{-1mm}
    \subfigure[]{
    \includegraphics[height=0.32\columnwidth]{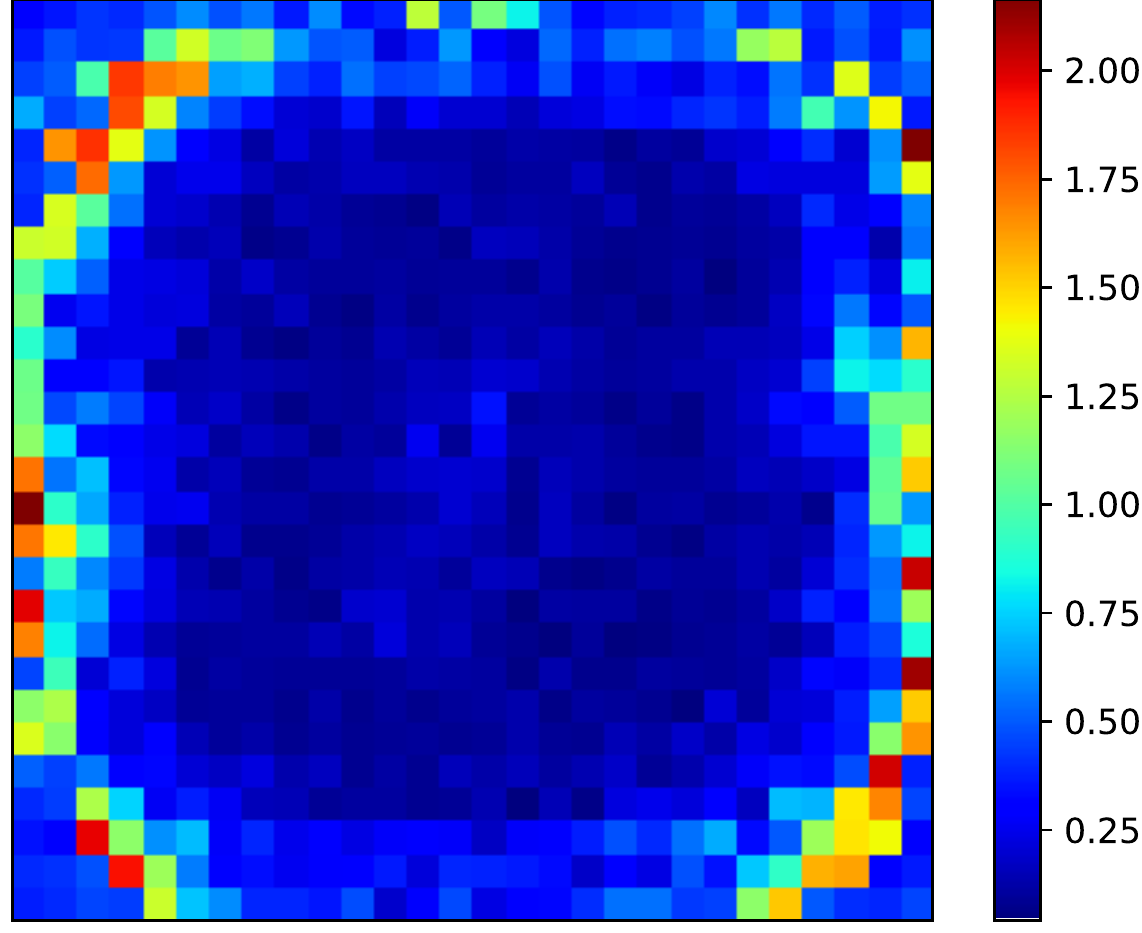}
    }
    \subfigure[]{
    \includegraphics[height=0.32\columnwidth]{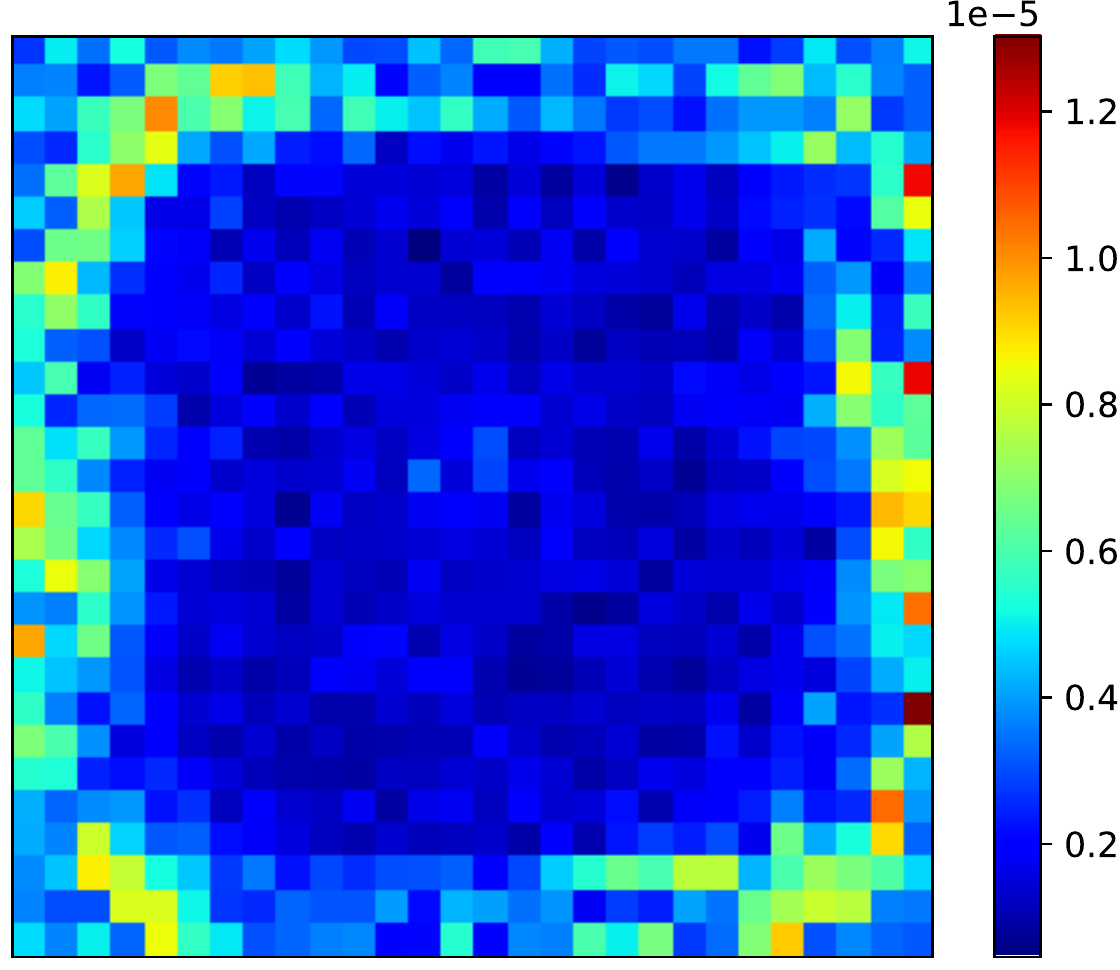}
    }
    \hspace{2mm}
    \subfigure[]{
    \includegraphics[height=0.32\columnwidth]{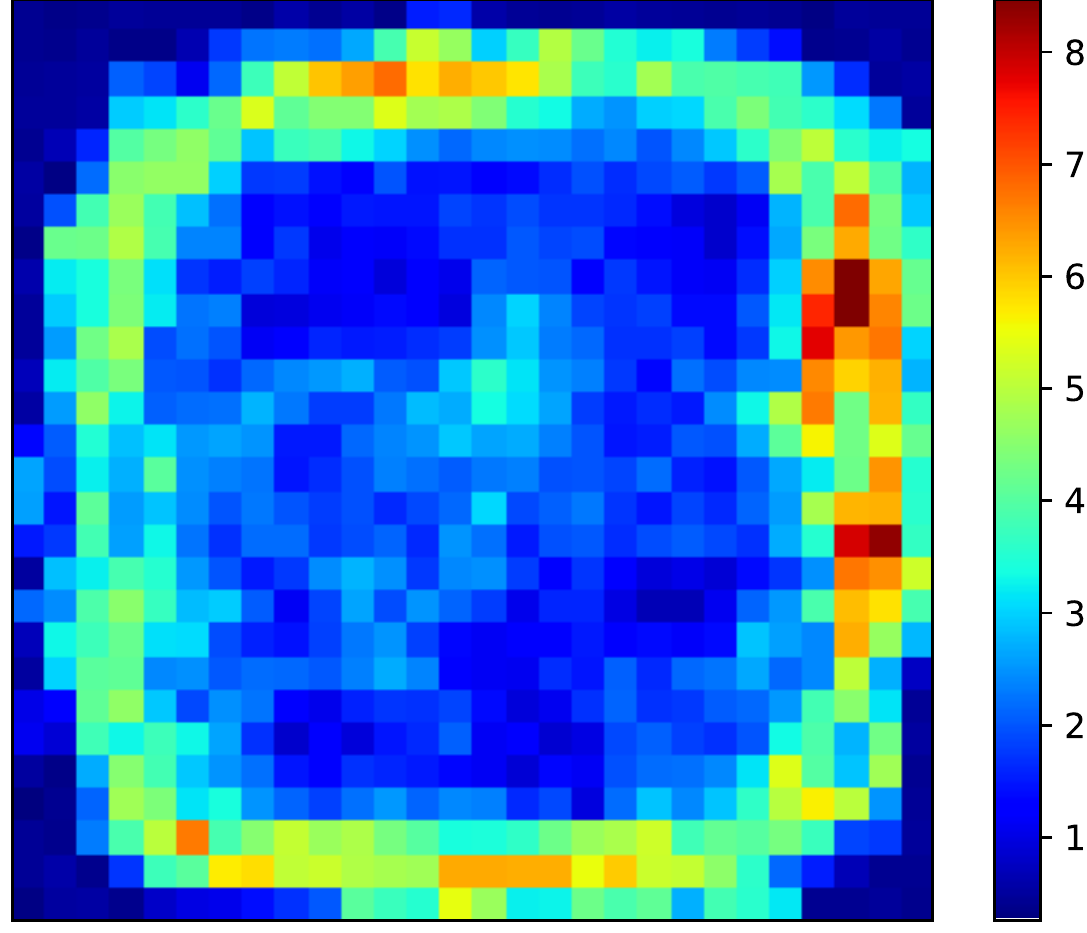}
    }
    \hspace{1mm}
    \subfigure[]{
    \includegraphics[height=0.32\columnwidth]{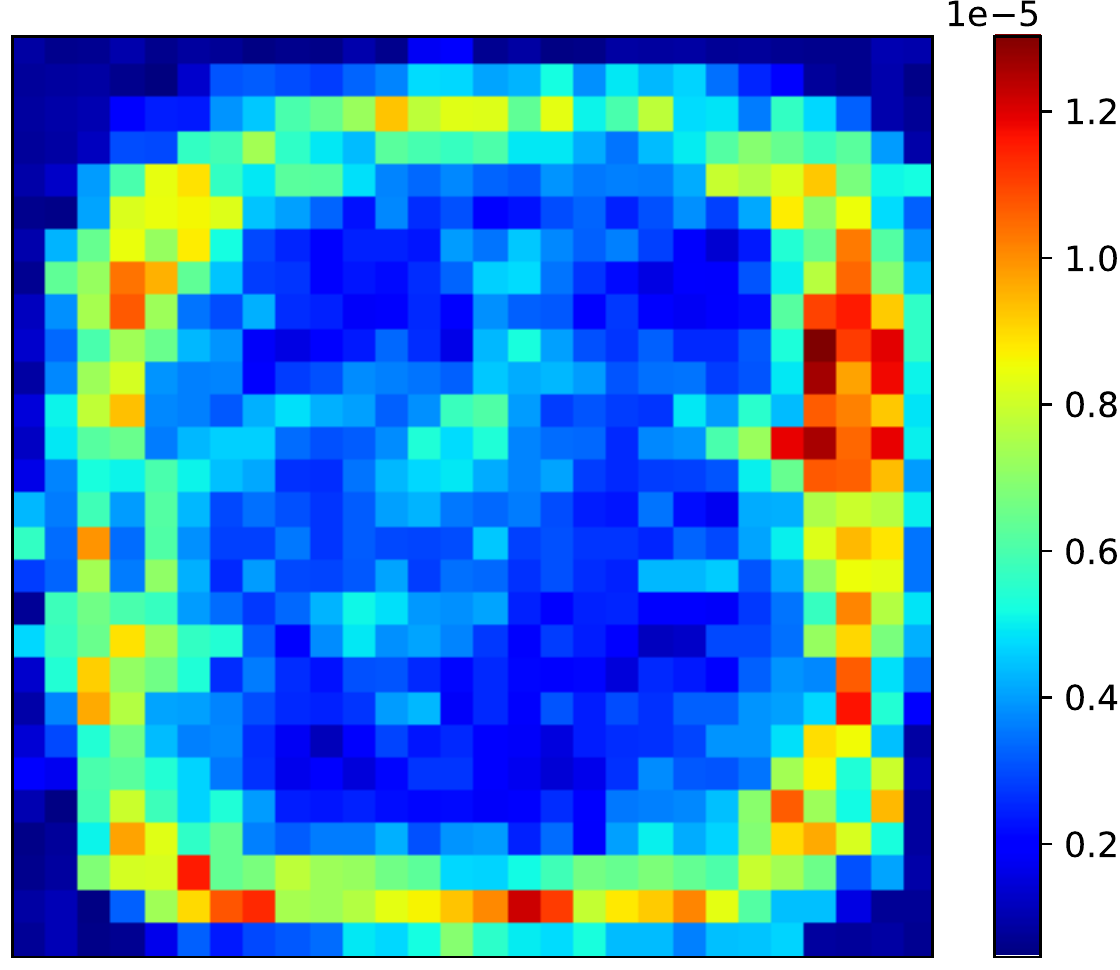}
    }
    \subfigure[]{
    \includegraphics[height=0.32\columnwidth]{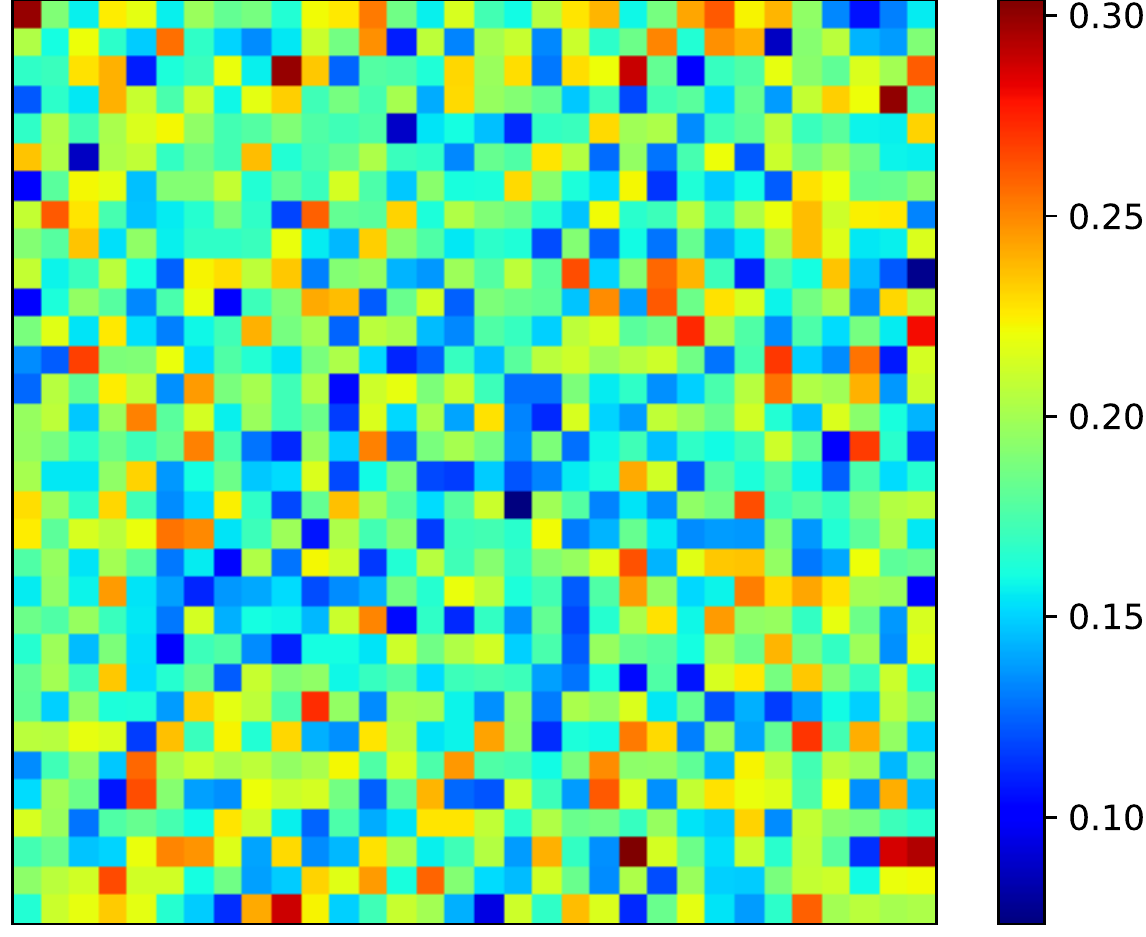}
    }
    \subfigure[]{
    \includegraphics[height=0.32\columnwidth]{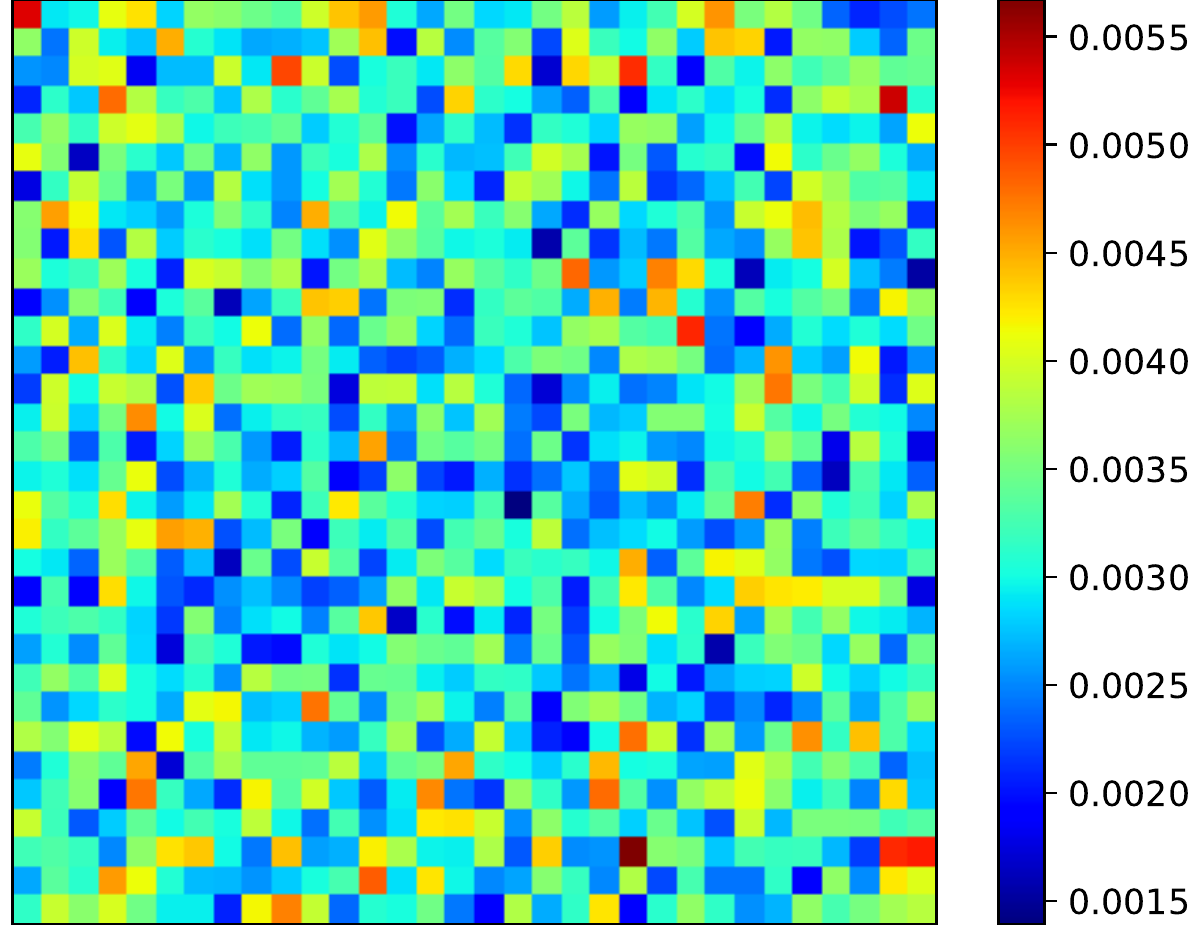}
    }
    \subfigure[]{
    \includegraphics[height=0.32\columnwidth]{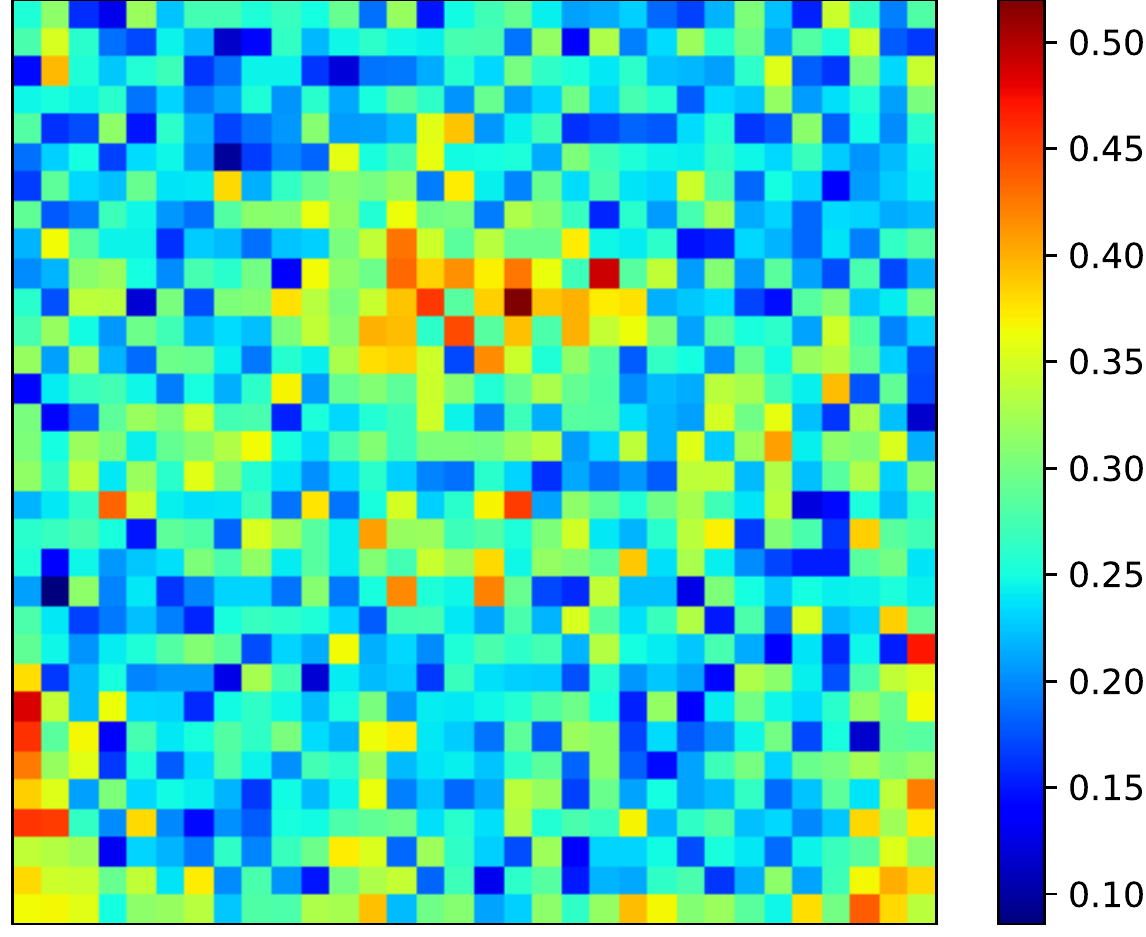}
    }
    \subfigure[]{
    \includegraphics[height=0.32\columnwidth]{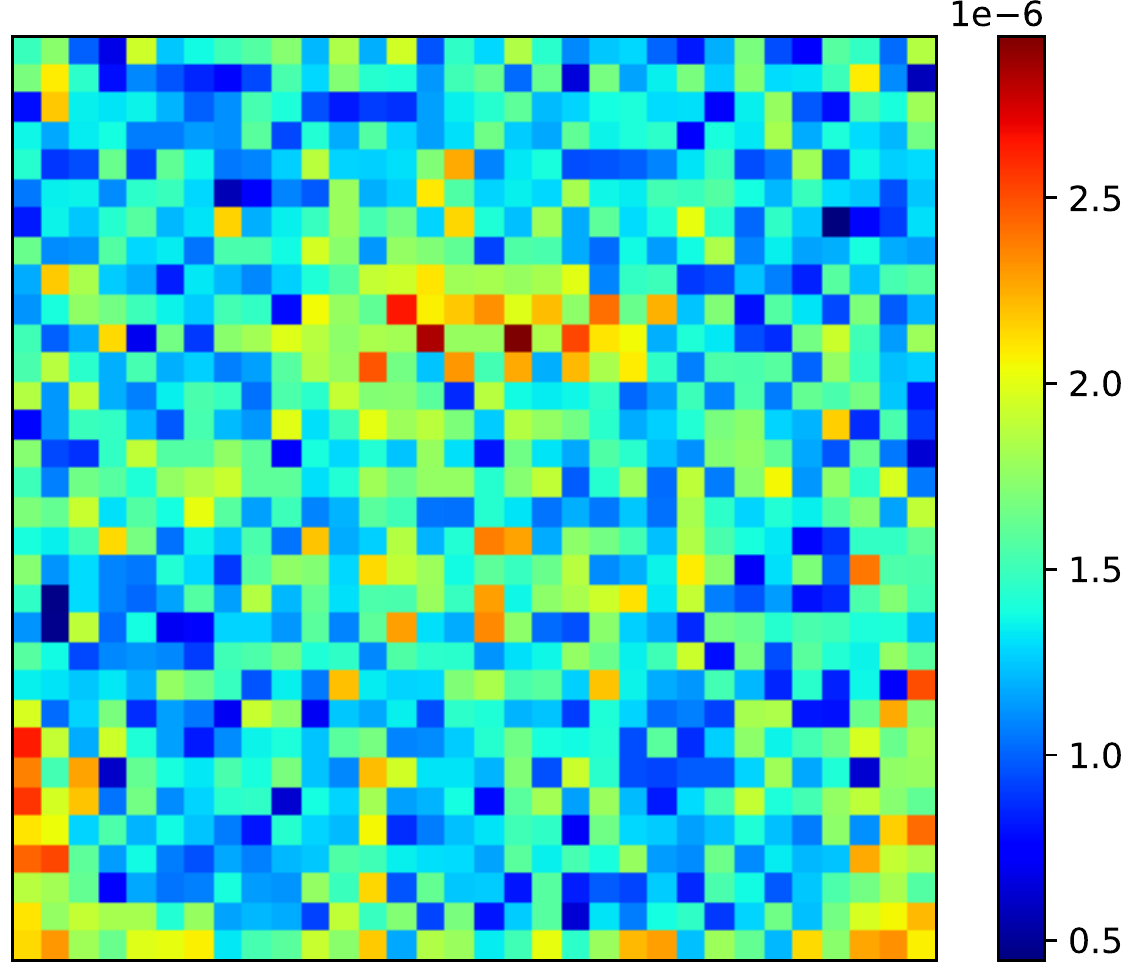}
    }
    \caption{Top Row:  Sensitivity (gradient of loss w.r.t. input) of each input for a single-layer neural network trained on the MNIST dataset with (a) no activation function (linear) with MSE loss and (c) softmax activation function with categorical crossentropy loss.  (b) and (d) show the corresponding 1-norms for each column vector in the 10$\times$784 weight matrix.  Bottom Row:   Sensitivity (gradient of loss w.r.t. input) of each input for a single-layer neural network trained on the CIFAR-10 dataset with (e) no activation function (linear) with MSE loss and (g) softmax activation function with categorical crossentropy loss.  (f) and (h) show the corresponding 1-norms for each column vector in the 10$\times$1024 weight matrix (corresponding to only the first color channel).}
    \label{fig:heat_map}
    \vspace{-4mm}
\end{figure*}

If the attacker does not have access to the crossbar output, then they must formulate an attack based only on knowledge of the 1-norms of each column of $\mathbf{W}$.  It stands to reason that the 1-norm of each weight matrix column has some relationship to the importance of the corresponding input feature.  To see this, consider the sensitivity of the loss function to the input of a single-layer neural network:
\begin{equation}
\frac{\partial \mathcal{L}}{\partial u_{j}}=\sum\limits_{i=1}^{M}\frac{\partial \mathcal{L}}{\partial \hat{y}_{i}}\frac{\partial \hat{y}_{i}}{\partial s_{i}}\frac{\partial s_{i}}{\partial u_{j}}=\sum\limits_{i=1}^{M}\frac{\partial \mathcal{L}}{\partial \hat{y}_{i}}f'(s_{i})w_{ij}.
\end{equation}
The slope of $f$ is non-negative for most popular activation functions (sigmoid, ReLU, etc.).  The sign of $\partial \mathcal{L}/\partial \hat{y}_{i}$ and $w_{ij}$ will depend on the loss function and the training data.  We can also write 
\begin{equation}
\left|\frac{\partial\mathcal{L}}{\partial u_{j}}\right|\le\sum\limits_{i=1}^{M}\left|\frac{\partial \mathcal{L}}{\partial \hat{y}_{i}}f'(s_{i}) \right| \left| w_{ij}\right|
\end{equation}
where equality will only hold when $\partial \mathcal{L}/\partial \hat{y}_{i}$ and $w_{ij}$ have the same sign (assuming that the slope of the activation function is non-negative).  However, this will not necessarily be the case.

Figure \ref{fig:heat_map} shows the correlation between the magnitude of the sensitivity of each input and 1-norms of the weight matrix columns for a 1-layer neural network trained on MNIST (top row) and CIFAR-10 (bottom row) datasets.  There are 4 cases:  MNIST with a linear output and mean squared error (MSE) loss (a and b), MNIST with a softmax output and crossentropy loss (c and d), CIFAR-10 with a linear output and MSE loss (e and f), and CIFAR-10 with a softmax output and crossentropy loss (g and h).  In each pair of figures, the one on the left is the magnitude of the sensitivity, averaged over the test set, and the one on the right show the 1-norms of the columns of the weight matrix.  Visually, one can observe a correlation between each pair, which is more pronounced for the MNIST dataset, but also present for the CIFAR-10 dataset.  This implies that, on average, the 1-norm is a good indicator of the mean sensitivity of the loss w.r.t. the inputs.  To quantify this, Table \ref{tab:corr} shows the correlation coefficients between these two metrics.  The right two columns correspond to the correlation of the mean sensitivity and the 1-norms averaged over 5 independent runs.  We see large correlations in all cases, which is in agreement with our visual inspection of Figure \ref{fig:heat_map}.  However, the average correlation coefficient between the sensitivity to each input and the 1-norms (left two columns of data in Table \ref{tab:corr}) is much lower.  This means that, while the 1-norms may provide a good summary about the average sensitivity, they will not be as effective in providing information about the sensitivity of individual input patterns. 

\begin{table}[!t]
\footnotesize
\centering
\caption{Correlation coefficients between the magnitude of the loss sensitivity and the 1-norms of the weight matrix columns for a 1-layer neural network on 2 different datasets.}
\begin{tabular}{llllllllll} 
\cmidrule[\heavyrulewidth]{1-6}
                          &            & \multicolumn{2}{l}{Mean Correlation} & \multicolumn{2}{l}{Correlation of Mean} &  &  &  &   \\
Dataset                   & Activation & Train & Test                         & Train & Test                            &  &  &  &   \\ 
\cmidrule{1-6}
\multirow{2}{*}{MNIST}    & Linear     & 0.70  & 0.70                         & 0.99  & 0.98                            &  &  &  &   \\
                          & Softmax    & 0.52  & 0.52                         & 0.92  & 0.92                            &  &  &  &   \\
\multirow{2}{*}{CIFAR-10} & Linear     & 0.26  & 0.26                         & 0.87  & 0.87                            &  &  &  &   \\
                          & Softmax    & 0.33  & 0.33                         & 0.91  & 0.91                            &  &  &  &   \\ 
\cmidrule[\heavyrulewidth]{1-6}
                          &            &       &                              &       &                                 &  &  &  &   \\
                          &            &       &                              &       &                                 &  &  &  &  
\end{tabular}
\label{tab:corr}
\vspace{-14mm}
\end{table}

\begin{figure*}[!t]
    \centering
    \subfigure[]{
        \includegraphics[width=0.7\columnwidth]{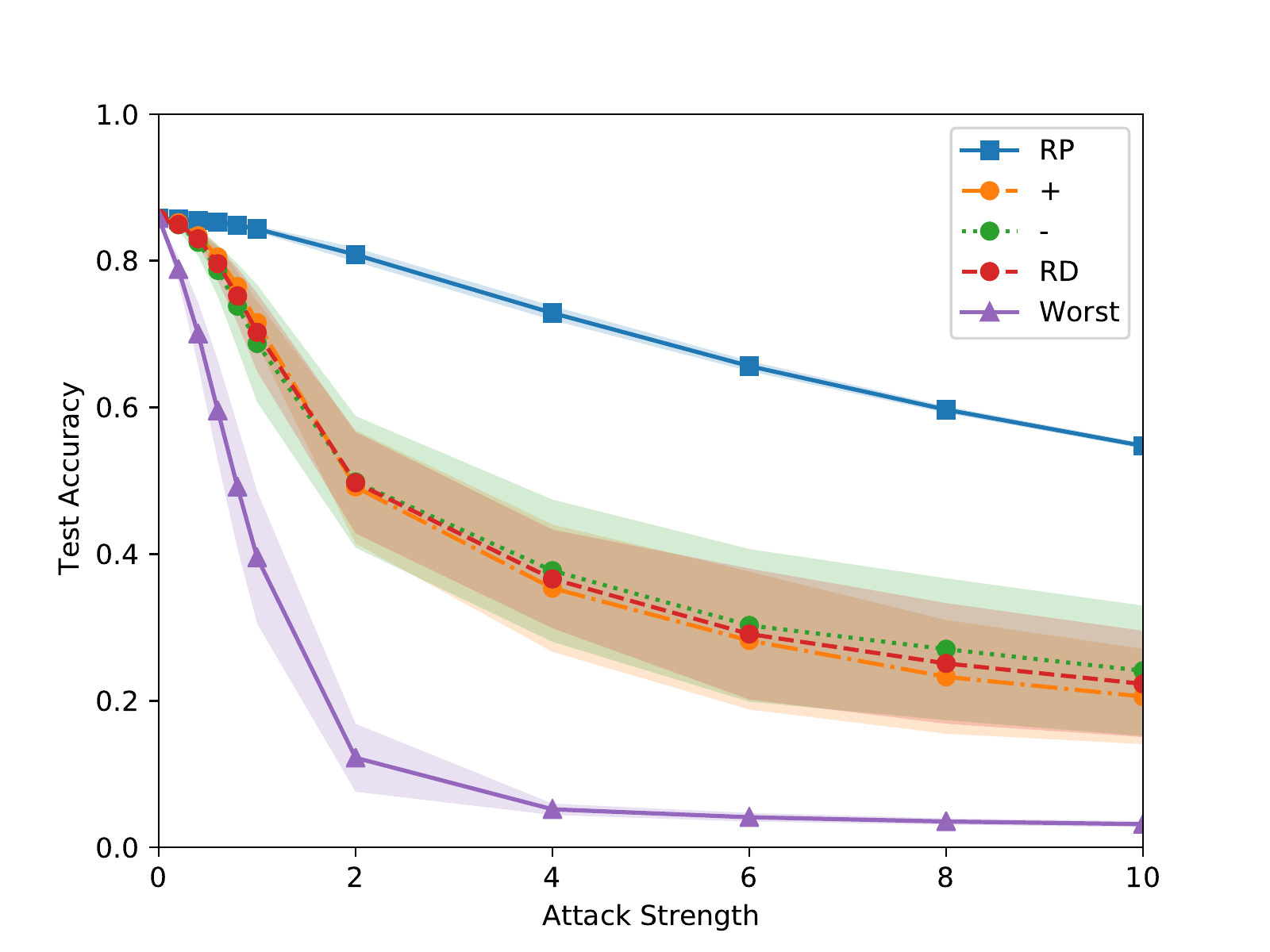}
    }
    \vspace{-2mm}
    \subfigure[]{
        \includegraphics[width=0.7\columnwidth]{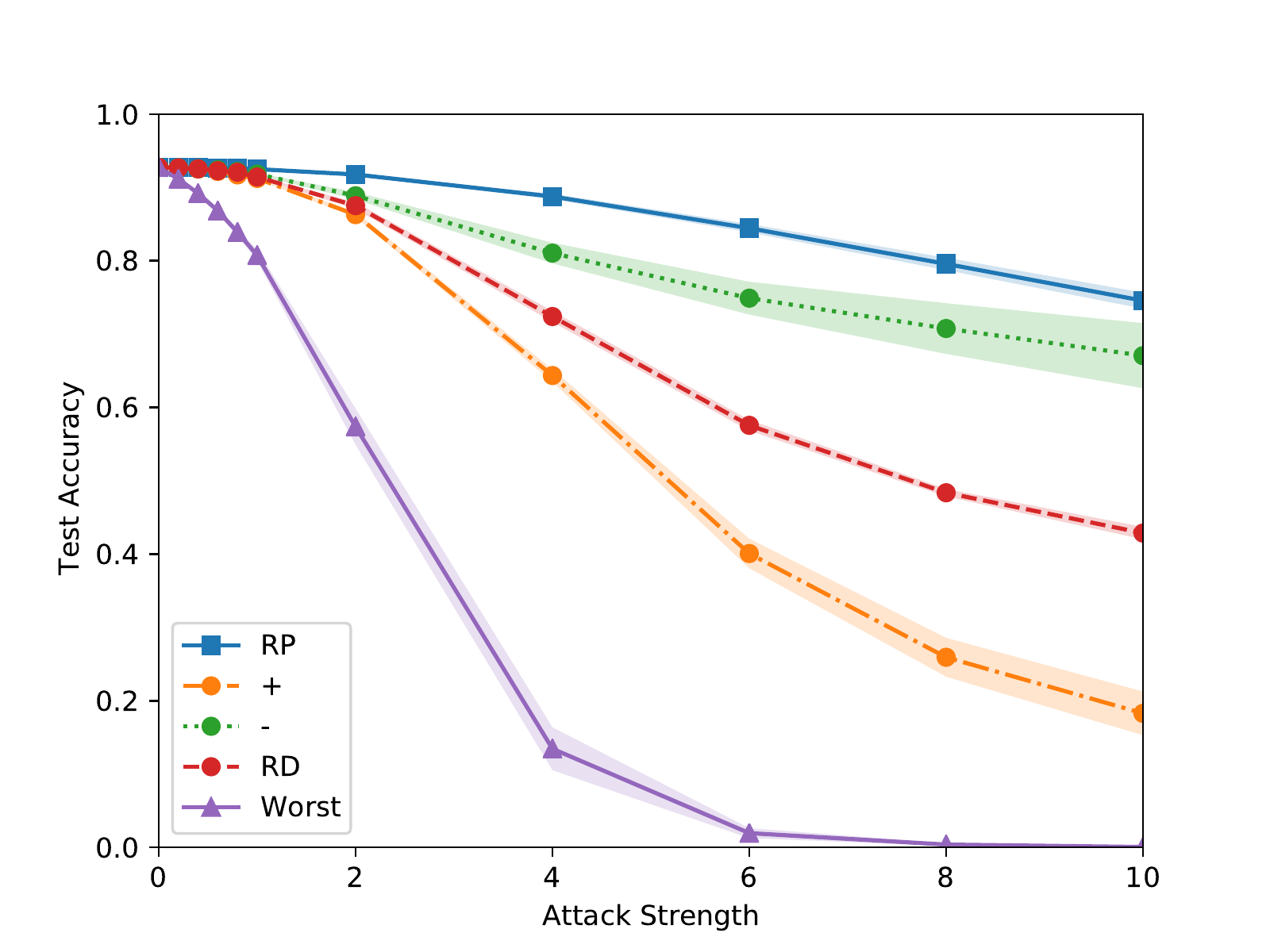}
    }
    \vspace{-2mm}
    \subfigure[]{
        \includegraphics[width=0.7\columnwidth]{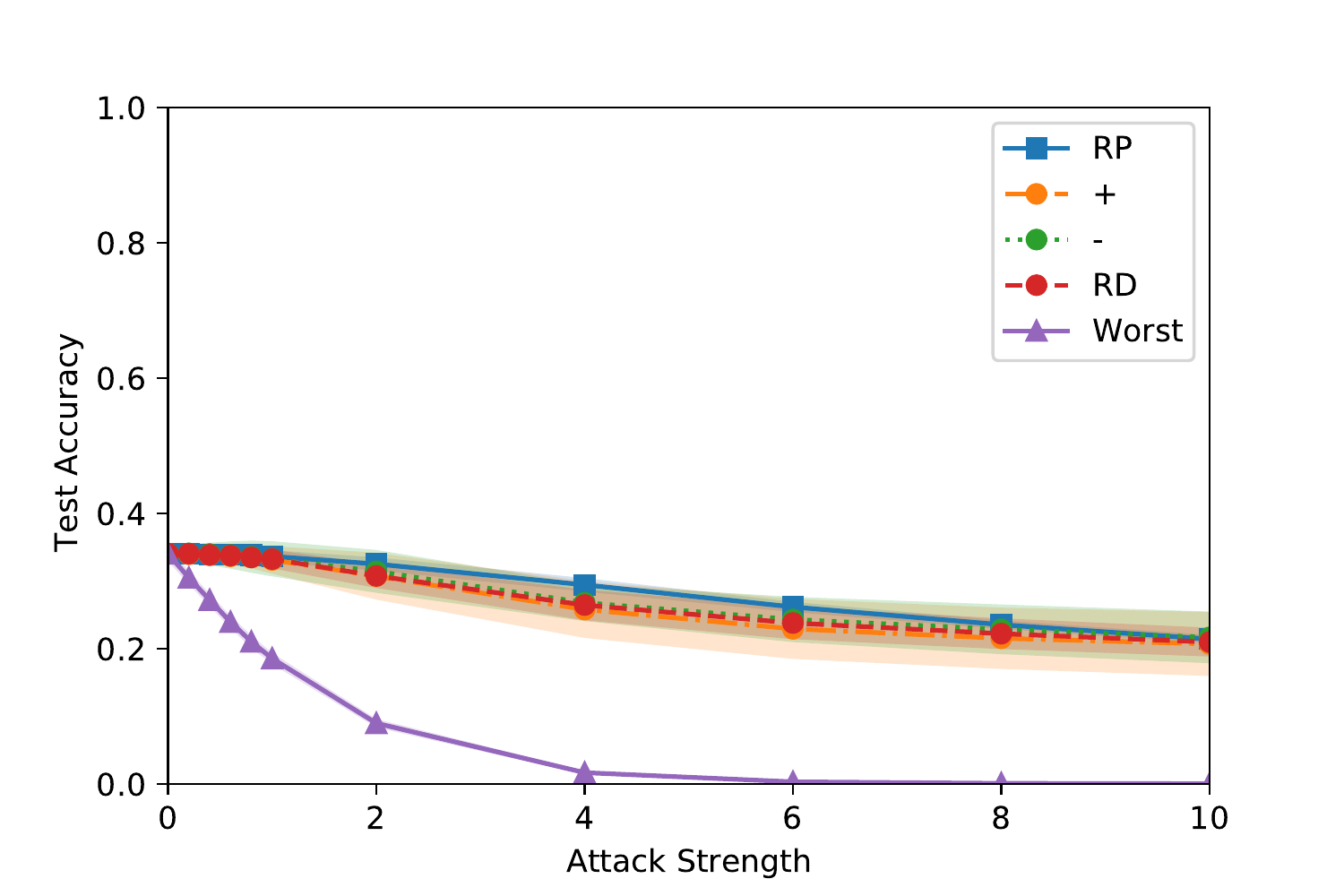}
    }
    \subfigure[]{
        \includegraphics[width=0.7\columnwidth]{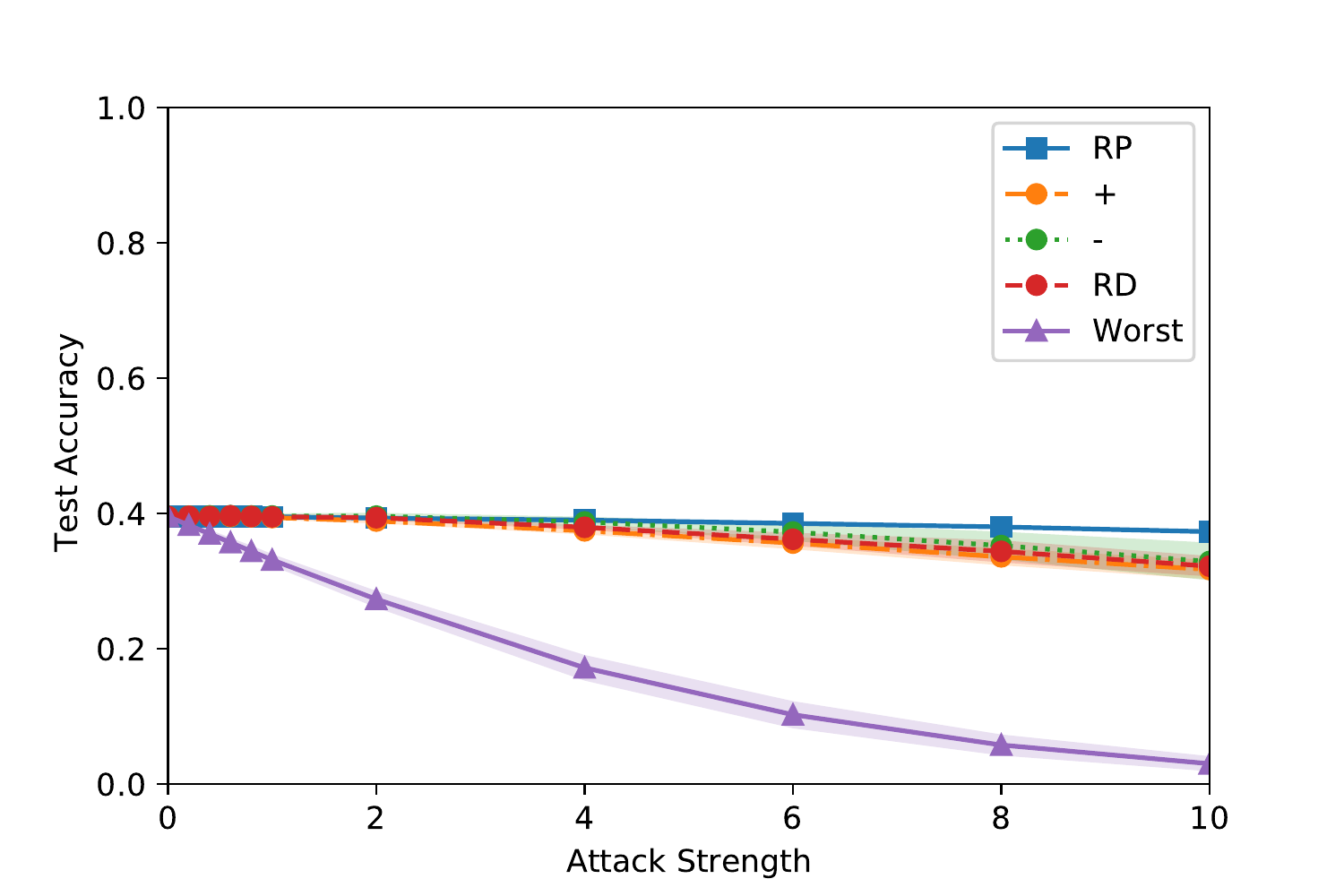}
    }
    \caption{Results of single-pixel attacks on 1-layer neural networks guided by power information.  (a) MNIST with linear output and MSE loss.  (b) MNIST with softmax output and crossentropy loss.  (c) CIFAR-10 with linear output and MSE loss.  (d) CIFAR-10 with softmax output and crossentropy loss.  In the legend, ``RP'' corresponds to a random pixel being modified with equal likelihood of adding or subtracting the attack strength.  Curves ``+'' and ``-'' correspond to attacking the pixel associated with the largest 1-norm of the weight matrix columns by adding or subtracting the attack strength, respectively.  ``RD'' corresponds to attacking the pixel associated with the largest 1-norm, but with an equal probability of either adding or subtracting the attack strength.  ``Worst'' corresponds to the worst-case, where the most sensitive pixel is perturbed in the direction of increasing loss.}
    \label{fig:spattacks}
    \vspace{-6mm}
\end{figure*}

Results of a single-pixel attack using the power information are shown in Figure \ref{fig:spattacks}.  The top two plots are for MNIST with a linear output and MSE loss (Figure \ref{fig:spattacks}(a)) and MNIST with a softmax output and crossentropy loss (Figure \ref{fig:spattacks}(b)), respectively.  The bottom two plots are for CIFAR-10 with a linear output and MSE loss (Figure \ref{fig:spattacks}(c)) and CIFAR-10 with a softmax output and crossentropy loss (Figure \ref{fig:spattacks}(d)), respectively.  Shown in each plot is test accuracy of the network when one of the input pixels is modified by either adding or subtracting the attack strength, with 5 different methods.  In the random pixel (RP) method, a random pixel is chosen to be modified for each image in the test set, with an equal probability of addition or subtraction.  This method is expected to be the least effective at reducing the training accuracy since it does not use any model information.  In fact, a hallmark of neural networks is that they are relatively robust to inputs with added random noise.  In the worst-case method (Worst), the pixel with the highest sensitivity, measured as the gradient of the loss w.r.t. the input, is modified in the direction of the loss gradient.  Essentially, this is an FGSM attack on a single pixel.  This is expected to give a lower bound for the test accuracy.  The other 3 methods use the 1-norms of the weight matrix columns to choose which pixel to attack.  The pixel associated with the largest 1-norm is chosen to add or subtract the attack strength.  We tested the case of adding only (+), subtracting only (-), and randomly choosing whether to add or subtract the attack strength (RD).  In all cases using 1-norm information improves the attack efficacy, with adding the attack strength being the most effective and subtracting being the least effective (most apparent in Figure \ref{fig:spattacks}(b)).  However, it is observed that the utility of the power information in crafting adversarial examples depends strongly on the dataset, the loss function, and type of output layer.  Note that we have also tested the proposed scheme with multiple-pixel attacks by adjusting the pixels associated with the top $N$ 1-norms.  In that case, we found that the attack success tends to decrease with $N$.  This is expected, since the probability of guessing the correct perturbation direction for all $N$ pixels is $(1/2)^{N}$.  

Finally, we note that the attack scheme presented above requires a measurement for each crossbar input to determine the 1-norm of every column.  It may be possible to estimate the location of the largest 1-norm with fewer queries using discrete optimization over the image locations.  However, the success of this strategy will depend on several factors such as smoothness of the 1-norm.  For example, it can be seen in Figure \ref{fig:spattacks} that the 1-norm data for the MNIST dataset has gradually changing values over the image location.  Therefore, it may be possible to employ some standard optimization techniques or search strategies (e.g. binary search).  On the other hand, the 1-norm for the CIFAR-10 dataset changes much more rapidly over image locations, which may cause difficulty in searching for the maximum.   

\section{Case 2:  Attacker Has Access to Output}
\label{section:output}

\begin{figure*}[!t]
\subfigure[]{
\centering
\includegraphics[width=0.32\textwidth]{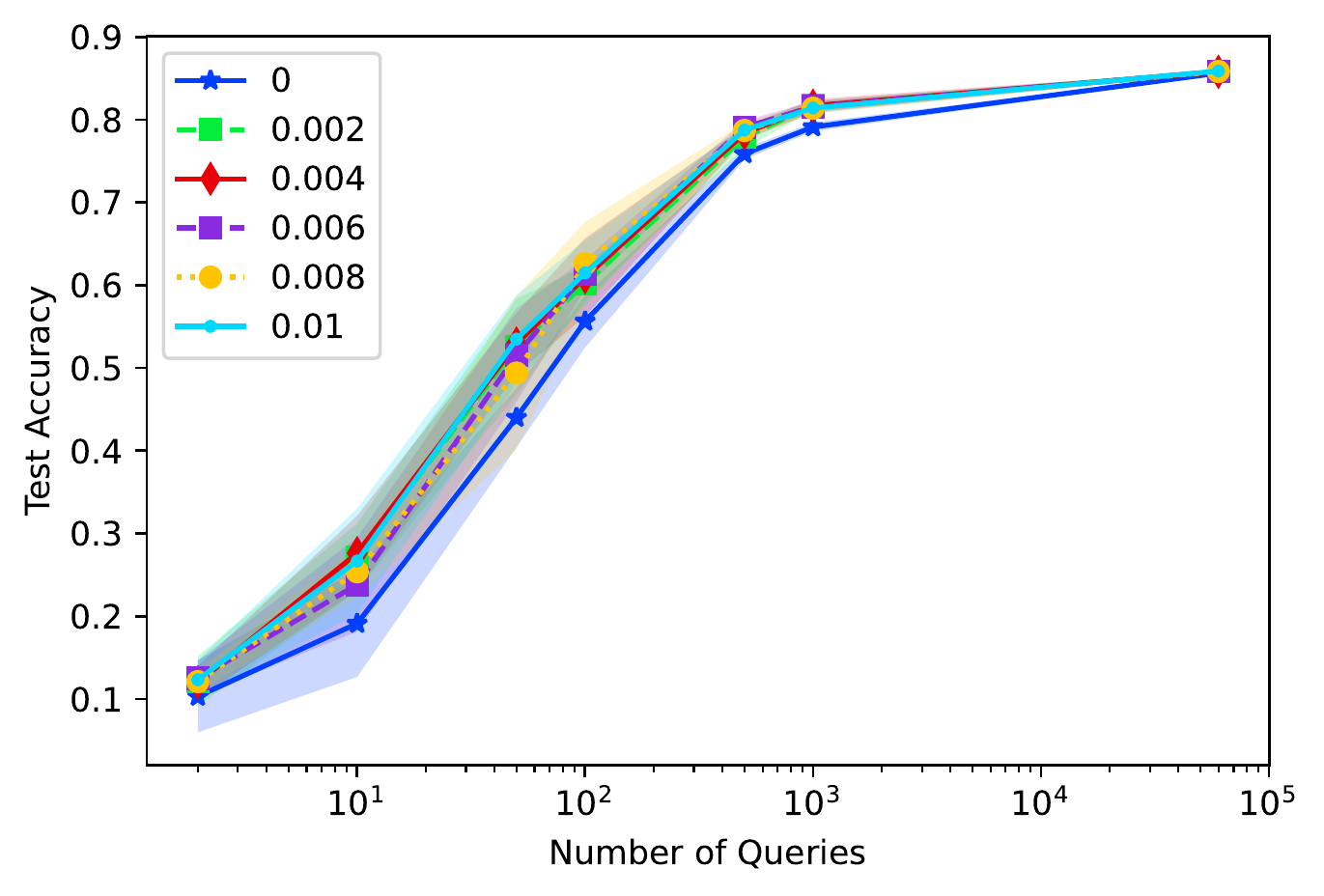}
}
\subfigure[]{
\centering
\includegraphics[width=0.32\textwidth]{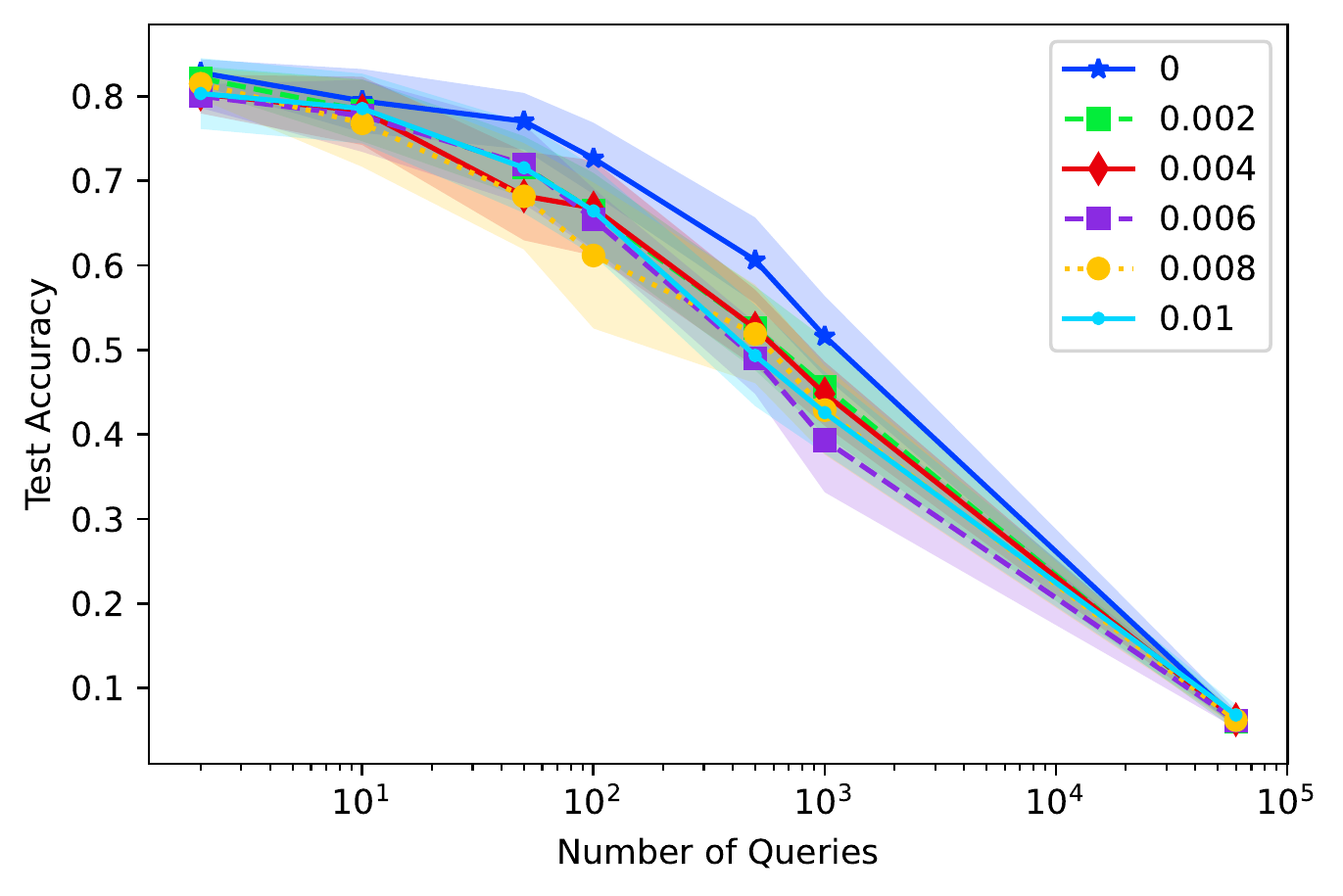}
}
\subfigure[]{
\centering
\includegraphics[width=0.32\textwidth]{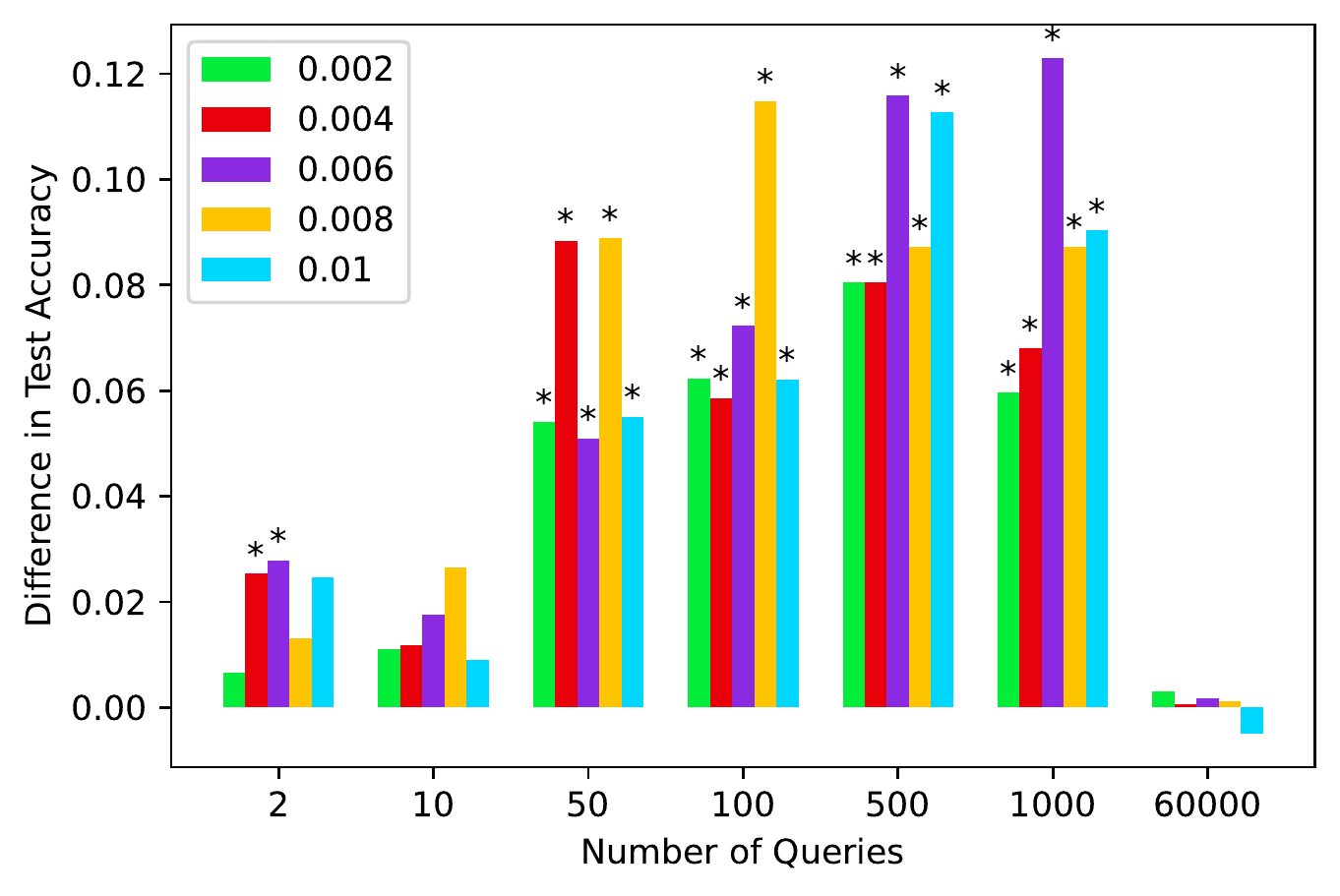}
}
\vspace{-7mm}
\end{figure*}

\begin{figure*}[!t]
\subfigure[]{
\centering
\includegraphics[width=0.32\textwidth]{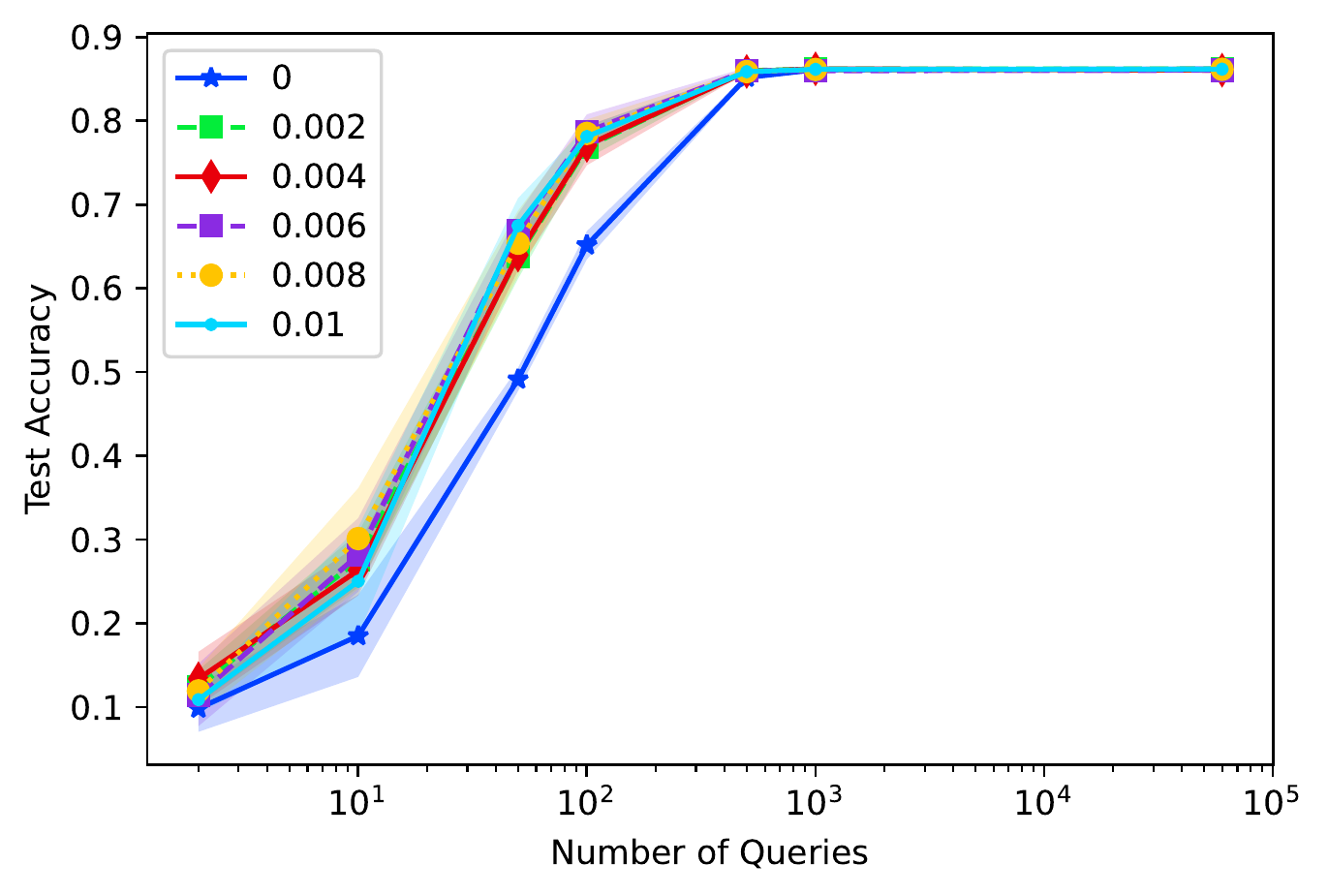}
}
\subfigure[]{
\centering
\includegraphics[width=0.32\textwidth]{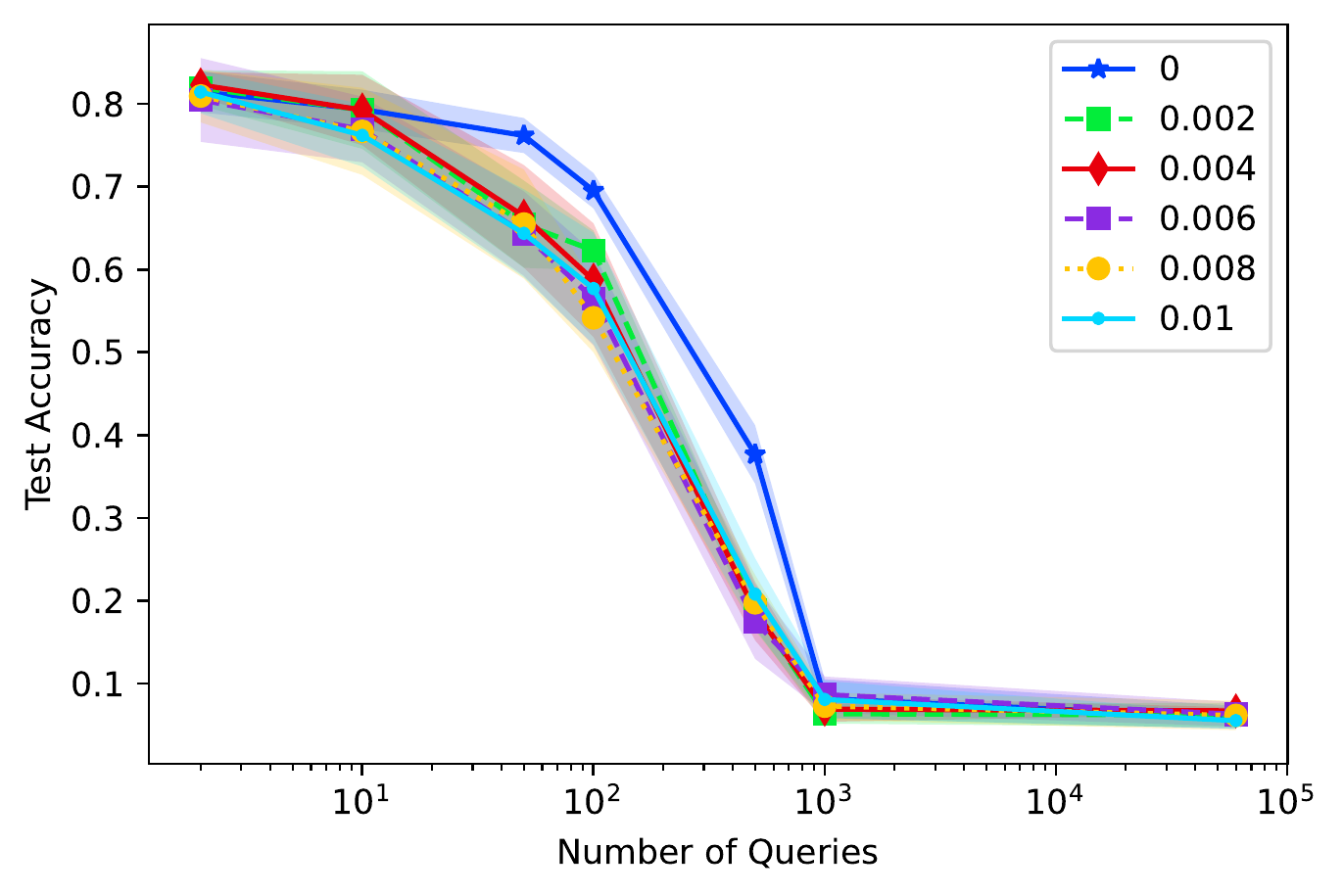}
}
\subfigure[]{
\centering
\includegraphics[width=0.32\textwidth]{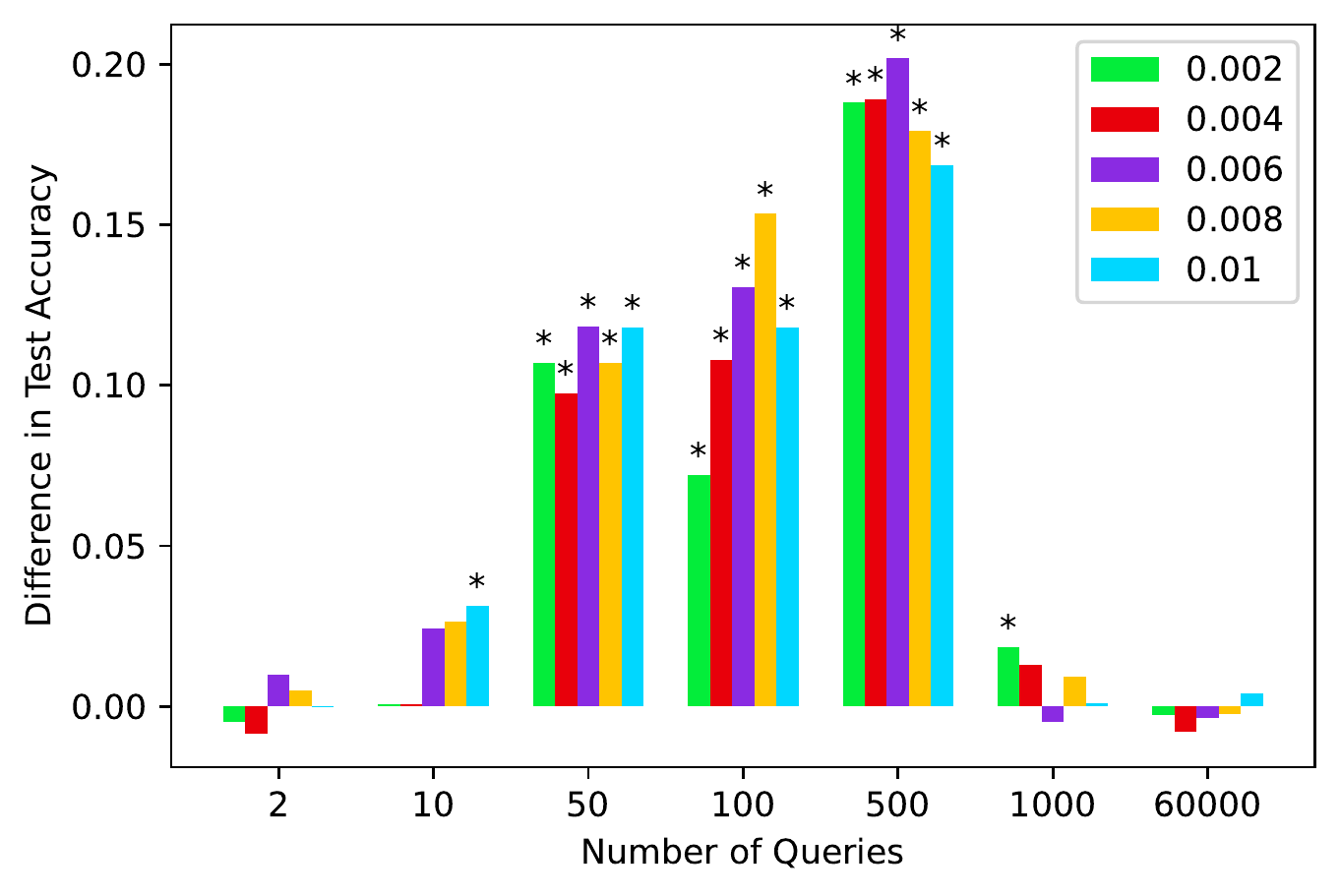}
}
\vspace{-6mm}
\end{figure*}

\begin{figure*}[!t]
\subfigure[]{
\centering
\includegraphics[width=0.32\textwidth]{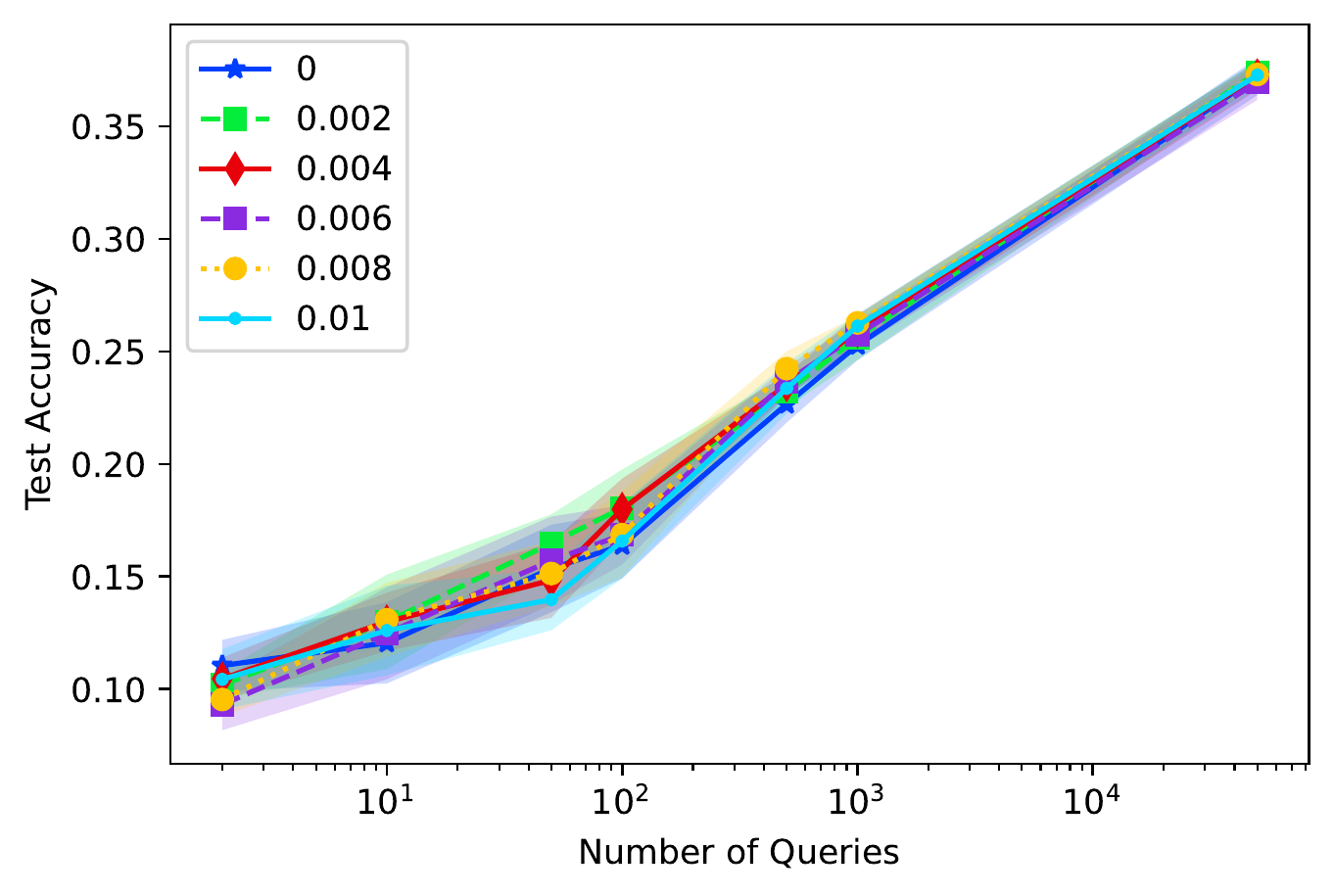}
}
\subfigure[]{
\centering
\includegraphics[width=0.32\textwidth]{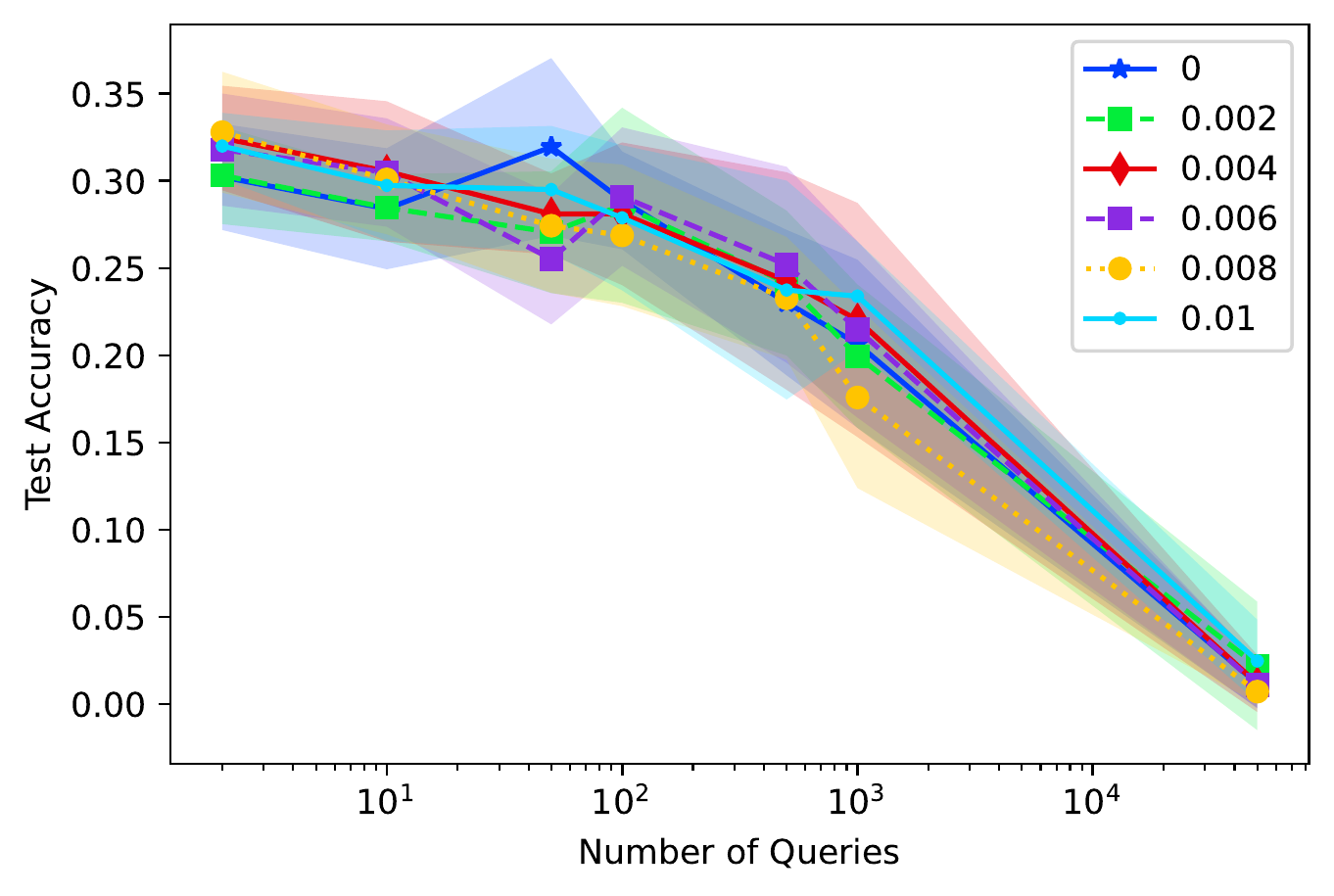}
}
\subfigure[]{
\centering
\includegraphics[width=0.32\textwidth]{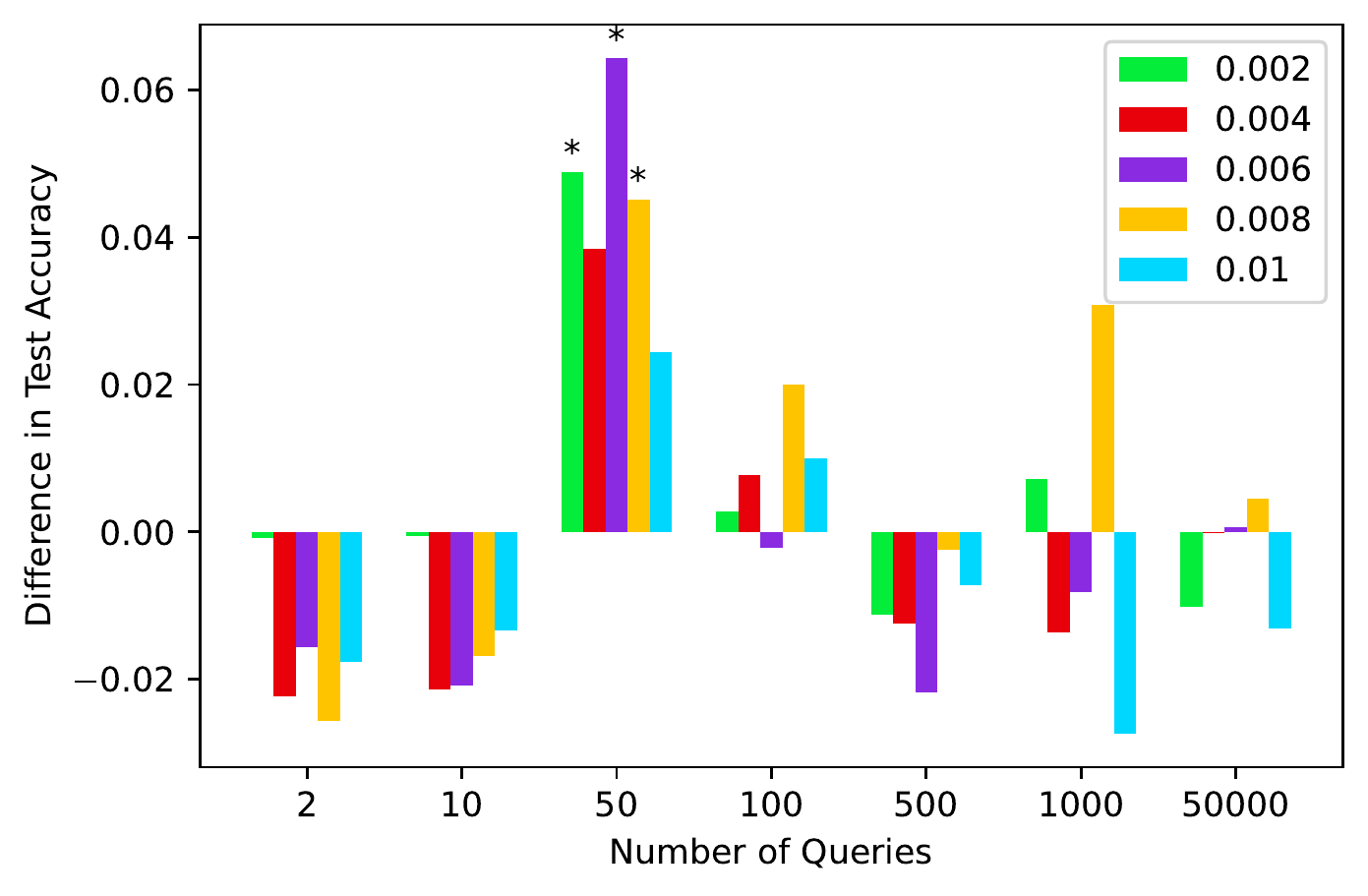}
}
\vspace{-6mm}
\end{figure*}

\begin{figure*}[!t]
\subfigure[]{
\centering
\includegraphics[width=0.32\textwidth]{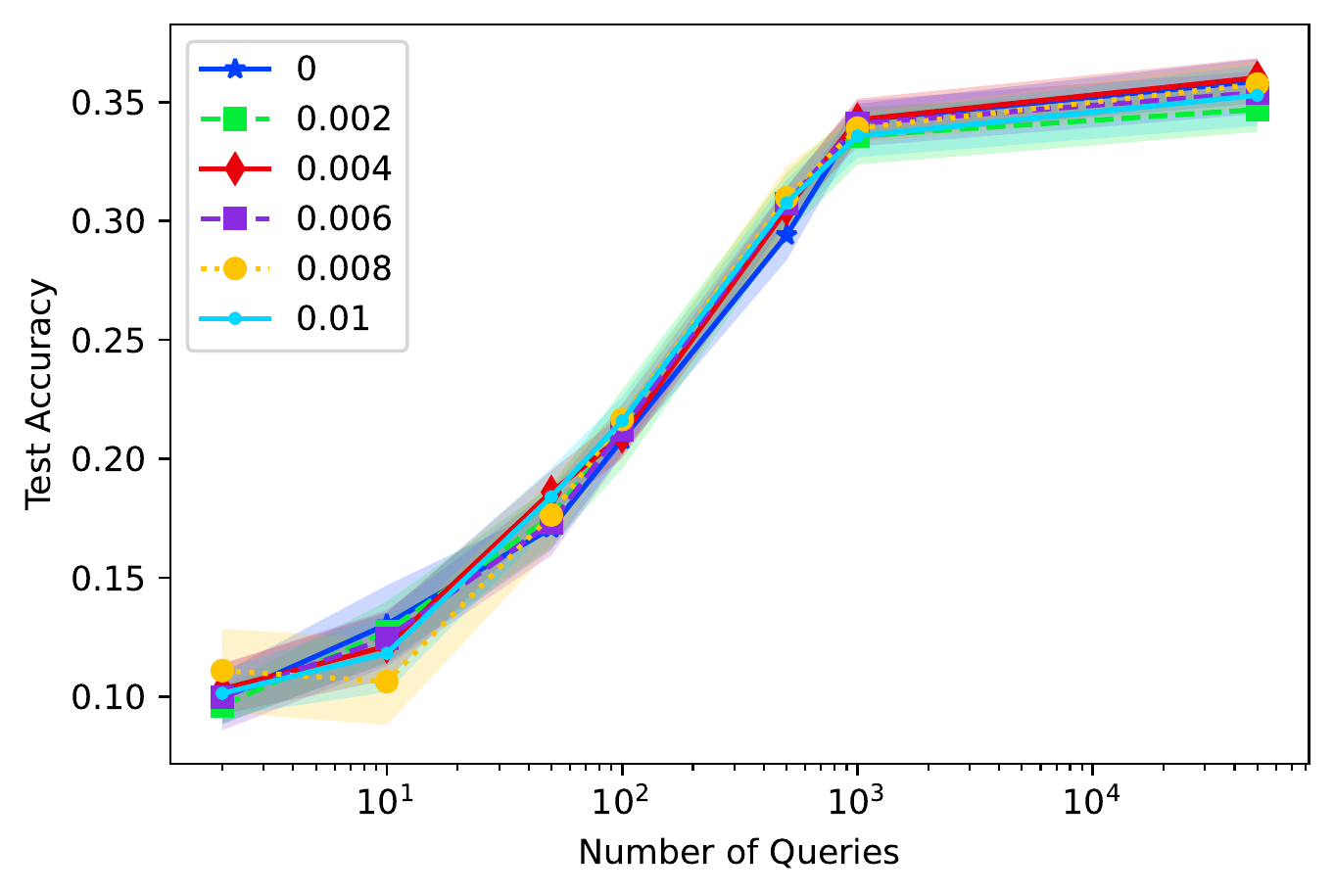}
}
\subfigure[]{
\centering
\includegraphics[width=0.32\textwidth]{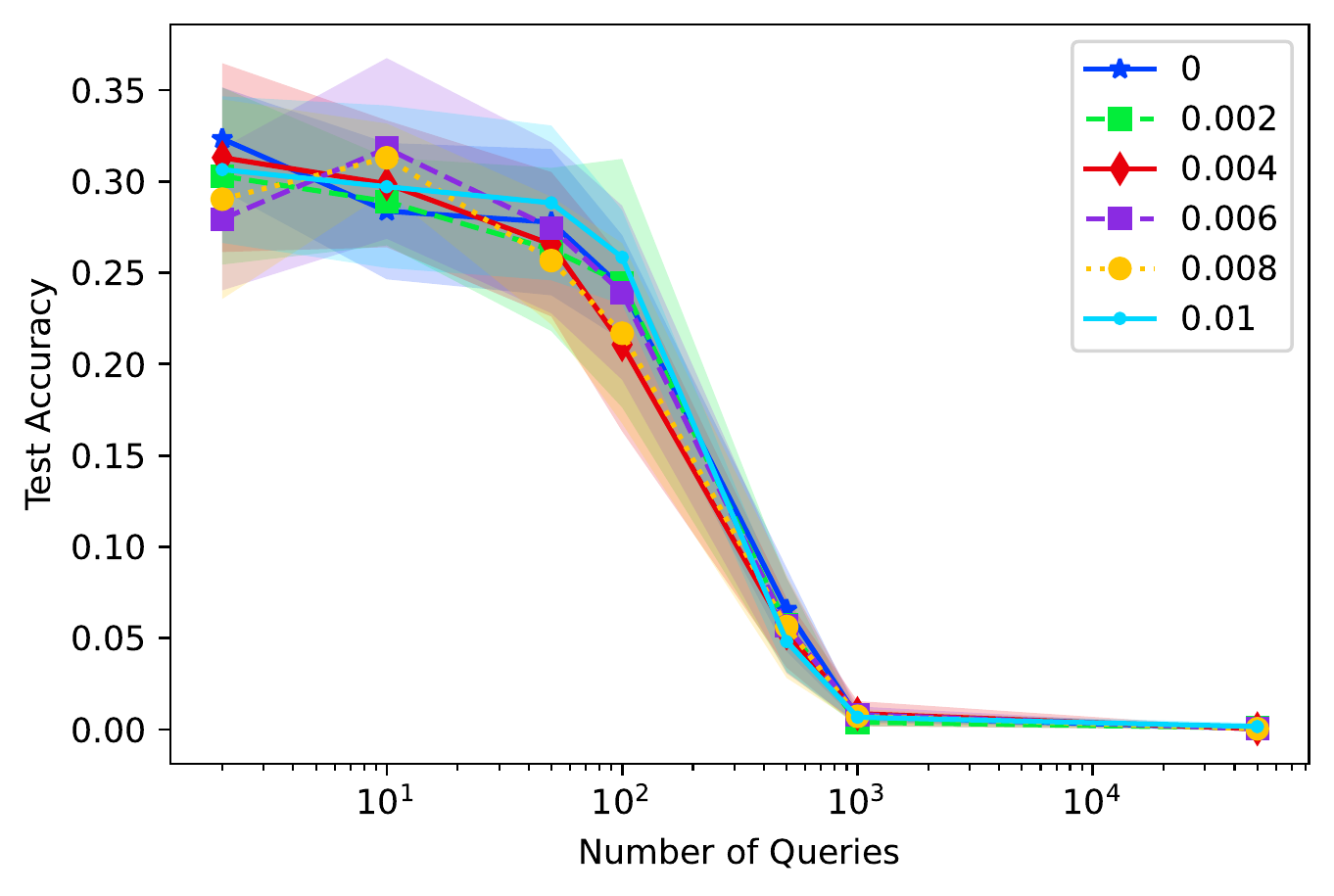}
}
\subfigure[]{
\centering
\includegraphics[width=0.32\textwidth]{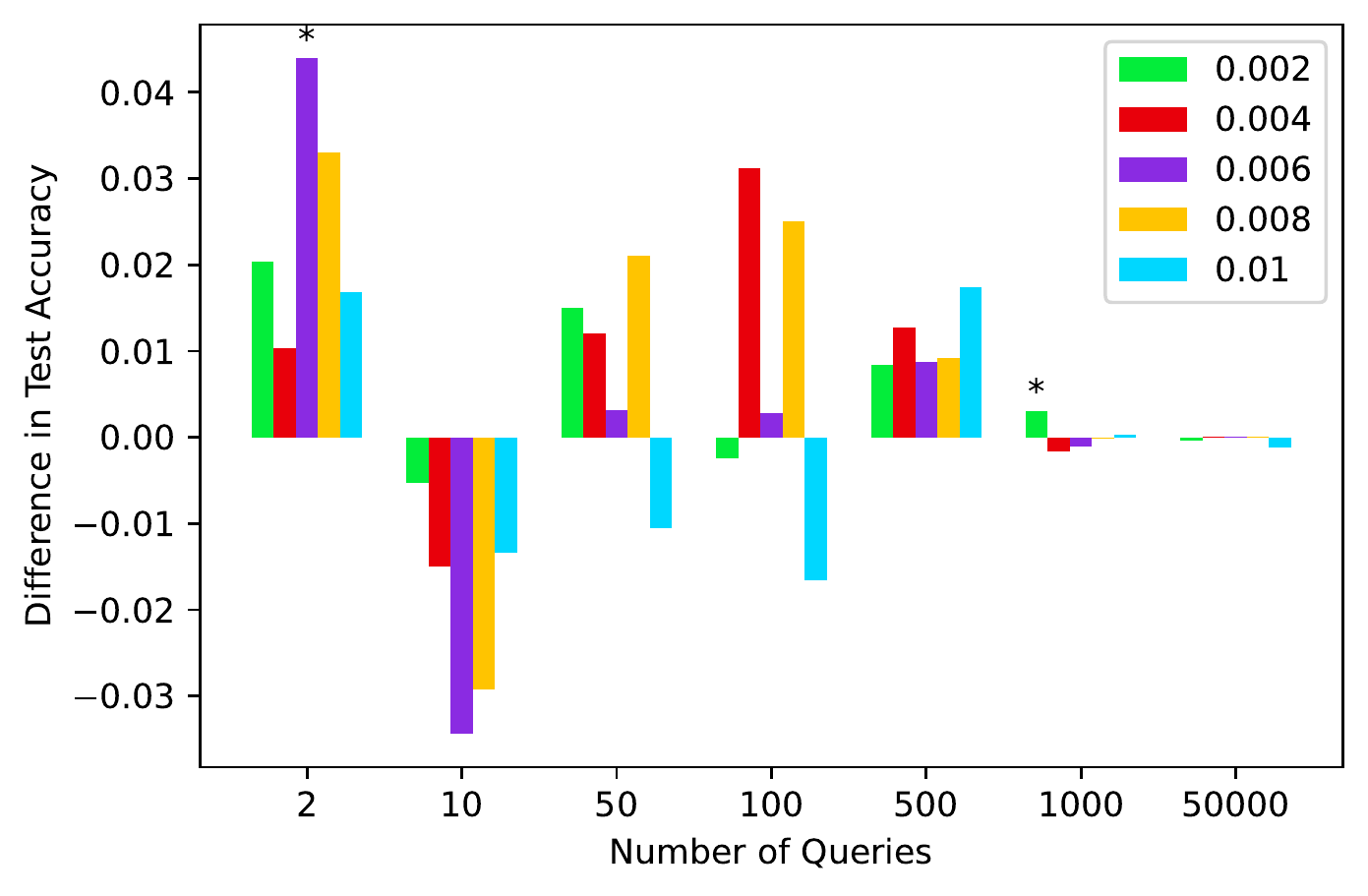}
}
\vspace{-6mm}
\caption{Incorporating power information in surrogate-based black box attacks.  Each row represents results from either a different dataset or different information extracted from the oracle.  \textbf{ROW 1}:  MNIST dataset with queries revealing only the label of the output.  \textbf{ROW 2}:  MNIST dataset with queries revealing the raw outputs.  \textbf{ROW 3}: CIFAR-10 dataset with  queries revealing only the label of the output.  \textbf{ROW 4}:  CIFAR-10 dataset with queries revealing the raw outputs. (a,d,g,j) Surrogate test accuracy vs. number of queries used to train the surrogate model for different power loss weight values (indicated in legend).  (b,e,h,k)  Oracle test accuracy for adversarial inputs produced using FGSM on the surrogate model with different numbers of queries.  (c,f,i,l)  Improvement in attack efficacy when power information is incorporated into training the surrogate model.  Asterisks indicate statistical significance with $p$-value $<$ 0.05.}
\label{fig:withoutputsresults}
\vspace{-3mm}
\end{figure*}

Now let us turn our attention to the case where the attacker has access to the output of the crossbar (after the activation function).  In the case of a linear activation function, we first observe that the power information will not be useful for input queries $\mathbf{u}=\beta\mathbf{e}_{j}$, where $\mathbf{e}_{j}$ is the vector such that the $j^{\text{th}}$ entry is 1 and the others are 0, and $\beta$ is a scalar.  This is because the output of the crossbar in that case would immediately reveal the $j^{\mathrm{th}}$ entries of the weight matrix since $\hat{\mathbf{y}}=\mathbf{W}\mathbf{u}=\mathbf{W}\beta\mathbf{e}_{j}$ can easily be solved for the $j^{\mathrm{th}}$ column of $\mathbf{W}$.  Also, assuming there is no measurement noise for crossbar current measurements, the power information is useless for the case where the queries are independent and the number of queries $Q$ is equal to or larger than the size of the input ($Q\ge N$).  Again, in this case, the system $\hat{\mathbf{y}}=\mathbf{W}\mathbf{u}$ is easily solvable as $\mathbf{W}={\mathbf{U}}^{\dagger}\hat{\mathbf{Y}}$, where $\dagger$ denotes the pseudoinverse, and $\mathbf{U}$ and $\hat{\mathbf{Y}}$ are matrices holding all of the query inputs and outputs, respectively.

With the exception of the two cases mentioned above, the crossbar power consumption information has the potential to reveal additional information about the model when combined with the crossbar outputs.  In this paper, we harness the power information by training a surrogate model using query information from the model that is being attacked (the ``oracle").  The oracles are identical to the single-layer neural networks presented in the last section and are queried by providing $Q$ inputs from the training set and then recording the crossbar output and power information.  Then, the query inputs, outputs, and power results are used as a dataset to train the surrogate model using the following loss function:
\begin{equation}
\mathcal{L}=\mathcal{L}_{out}+\lambda\mathcal{L}_{power}
\end{equation}
\noindent where $\mathcal{L}_{out}$ is the MSE of the outputs (only linear activation function is used), $\mathcal{L}_{power}$ is the MSE loss between the oracle and surrogate power information, and $\lambda$ is the power loss weight, which is a hyperparameter that determines the importance of the power information.  Figure \ref{fig:withoutputsresults} shows the results for MNIST (ROW 1 and ROW 2) and CIFAR-10 (ROW 3 and ROW 4) training sets.  Through experimentation, we found that smaller power loss weights (shown in the legends) tended to lead to less overfitting of the power data, leading to better results.  In this case, the power loss weight ranged from 0 to 0.01.  The results shown are averaged over 10 independent simulation runs with shaded error bars indicating the standard deviation.  In ROW 1 (MNIST) and ROW 3 (CIFAR-10), only the label of the oracle output is used, which represents the scenario where an attacker does not have access to the raw vector of outputs.  The label is equal to the output number with the largest value.  In contrast, ROW 2 (MNIST) and ROW 4 (CIFAR-10) show the results when the attacker has access to the raw output.  Within each row, the left plot shows the accuracy of the surrogate model on the test set vs. size of the training set used, which is obtained by querying the oracle.  The center plot shows the results of using the surrogate to generate FGSM adversarial inputs to the oracle with attack strength of 0.1.  Note that qualitatively similar trends are observed with other attack strengths as well.  The vertical axis is the adversarial test accuracy of the oracle, and the horizontal axis is the number of queries used to train the surrogate model.  The right plot shows the difference in the oracle model's accuracy degradation for the case where power information is and isn't used in the surrogate model.  Asterisks indicate statistically-significant results ($p$-value $<$ 0.05), determined by a Student's t-test.

For MNIST, notice that in all cases, the power consumption information improves the accuracy of the surrogate model on the test set and also improves query efficiency.  This means that the same level of oracle accuracy degradation can be achieved with fewer queries.  In addition, most of the improvements in the attack effectiveness are statistically significant.   Observe that the power information is most useful for moderate query sizes and does not improve the attack efficacy  when the query size becomes large.  This is expected behavior, as we have discussed above. In the case where the attacker has access to the raw outputs (ROW 2) there is significant increase in the attack effectiveness when the power consumption data are used (up to $\approx$ 20\%).  This improvement over the case where only labels are used can be attributed to the fact that only using label information essentially amounts to adding noise to the target outputs of the surrogate dataset, which will increase overfitting.  One can also observe in this case that the utility of the power information sharply drops off as the number of the queries becomes greater than the size of the model (784 for MNIST).  This is demonstrated by the flatness of the curves in Figures \ref{fig:withoutputsresults} (d) and (e) for large query sizes. 

Finally, we turn out attention to the case of the CIFAR-10 dataset (ROW 3 and ROW 4 in Figure \ref{fig:withoutputsresults}).  In both the case where only label data is collected from the queries (ROW 3) and the case where the raw outputs are revealed (ROW 4), there is a marked difference in the utility of the power information from the MNIST case.  While some statistically-significant improvements in the attack efficacy are achieved using the power data (up to $\approx$ 6\%), most of the results indicate no statistically-significant difference in the results with or without the power information.  This is also consistent with the results from Section \ref{section:nooutput}, which showed that power information is less useful for CIFAR-10 when the neural network outputs are unknown.  Future work should be conducted to gain a better understanding of this discrepancy, but we believe a possible reason is the low initial accuracy of the CIFAR-10 dataset, which is expected since we are only using a single-layer neural network.  Another reason may be the relative complexity of the CIFAR-10 dataset compared to MNIST and their different underlying statistics.  However, further study will be needed to verify these hypotheses.

\section{Conclusion}
\label{section:conclusion}

This paper provides the first investigation of adversarial attacks on NVM crossbar-based neural networks that incorporate power analysis.  While the results are theoretical and deal only with ideal crossbar behavior, we believe that they represent a critical step towards further research in this area.  Specifically, our results show that power information alone can reveal the sensitivity of a neural network's loss with respect to different inputs.  We also find that incorporating power information into the loss function of a surrogate model yields improved query efficiency for black box attacks.  Avenues for future work are numerous.  First, it will be important to consider non-ideal behavior of NVM crossbars and perform SPICE simulations to compare with the theoretical results presented here.  In addition, this work should be expanded to multi-layer neural networks in order to study the role of power information when attacking state-of-the-art deep neural networks.

\section*{Acknowledgments}

This material is based on research sponsored by the Air Force Research Laboratory under agreement number FA8750-20-2-0503. The U.S. Government is authorized to reproduce and distribute reprints for Governmental purposes notwithstanding any copyright notation hereon.  The views and conclusions contained herein are those of the authors and should not be interpreted as necessarily representing the official policies or endorsements, either expressed or implied, of the Air Force Research Laboratory or the U.S. Government.




\bibliographystyle{IEEEtran}
\footnotesize

\bstctlcite{IEEEexample:BSTcontrol}
\bibliography{refs.bib}

\end{document}